\documentclass{article}
\usepackage{ijcai23}

\usepackage{times}
\usepackage{soul}
\usepackage{url}
\usepackage[hidelinks]{hyperref}
\usepackage[utf8]{inputenc}
\usepackage[small]{caption}
\usepackage{graphicx}
\usepackage{amsmath}
\usepackage{amsthm}
\usepackage{amssymb}
\usepackage{booktabs}
\usepackage{algorithm}
\usepackage{algorithmic}
\usepackage{multicol}
\usepackage{multirow}
\usepackage{bm}
\usepackage{subfigure}
\usepackage[switch]{lineno}

\title{AllMatch: Exploiting All Unlabeled Data for Semi-Supervised Learning}

\author{
Zhiyu Wu
\and
Jinshi Cui\thanks{Corresponding author}
\affiliations
National Key Laboratory of General Artificial Intelligence, \\
School of Intelligence Science and Technology, Peking University\\
\emails
wuzhiyu@pku.edu.cn,
cjs@cis.pku.edu.cn
}

\begin{document}

\maketitle

\begin{abstract}
Existing semi-supervised learning algorithms adopt pseudo-labeling and consistency regulation techniques to introduce supervision signals for unlabeled samples. To overcome the inherent limitation of threshold-based pseudo-labeling, prior studies have attempted to align the confidence threshold with the evolving learning status of the model, which is estimated through the predictions made on the unlabeled data. In this paper, we further reveal that classifier weights can reflect the differentiated learning status across categories and consequently propose a class-specific adaptive threshold mechanism. Additionally, considering that even the optimal threshold scheme cannot resolve the problem of discarding unlabeled samples, a binary classification consistency regulation approach is designed to distinguish candidate classes from negative options for all unlabeled samples. By combining the above strategies, we present a novel SSL algorithm named AllMatch, which achieves improved pseudo-label accuracy and a 100\% utilization ratio for the unlabeled data. We extensively evaluate our approach on multiple benchmarks, encompassing both balanced and imbalanced settings. The results demonstrate that AllMatch consistently outperforms existing state-of-the-art methods.
\end{abstract}


\section{Introduction}
Semi-supervised learning (SSL)~\cite{zhu2005semi,rosenberg2005semi,berthelot2019mixmatch,sohn2020fixmatch}, a research topic that aims to boost the model's generalization performance by leveraging the potential of unlabeled data, has received extensive attention in recent years. Among the proposed techniques, the combination of pseudo-labeling~\cite{lee2013pseudo,arazo2020pseudo} and consistency regulation~\cite{sajjadi2016regularization,laine2016temporal}, as introduced by FixMatch~\cite{sohn2020fixmatch}, has emerged as a predominant approach. Specifically, FixMatch first assigns a pseudo-label to each unlabeled sample based on the prediction of its weakly augmented view. Subsequently, pseudo-labels exceeding a predefined confidence threshold are used as supervision for corresponding strongly augmented views, while those below the threshold are discarded. To ensure high-quality pseudo-labels, FixMatch uses a high constant threshold throughout training. However, this strategy results in the underutilization of unlabeled data, presenting a central challenge in SSL: making efficient use of unlabeled data.

\begin{figure}[t]
    \centering
    \subfigure[Dropped pseudo-label \textit{acc}.]{
        \includegraphics[width=0.22\textwidth]{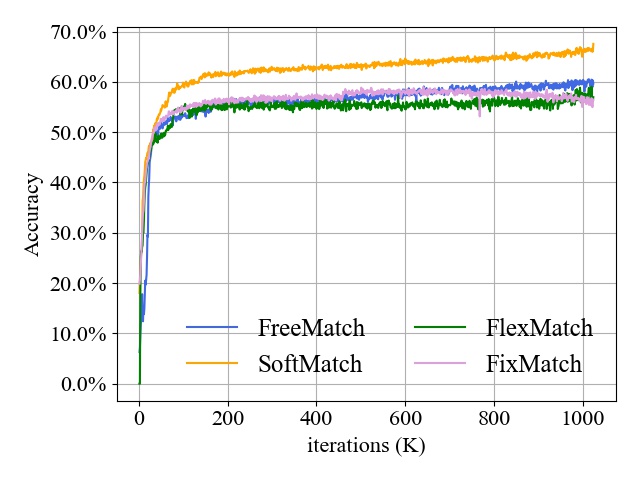}
        \label{dropped_pseudo_acc}
    }
    \subfigure[Top-5 \textit{acc} of pseudo-labels. ]{
        \includegraphics[width=0.22\textwidth]{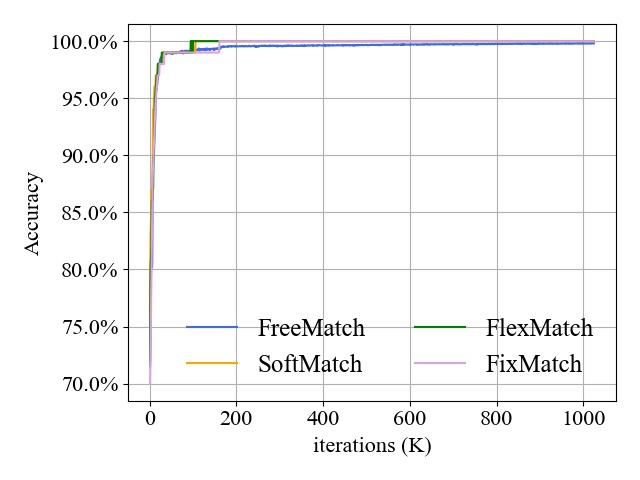}
        \label{top5_pseudo_acc}
    }
    \vspace{-0.3cm}
    \caption{Pilot study for pseudo-label quality on CIFAR-10 with 40 labels. Regarding SoftMatch, pseudo-labels with a confidence lower than $\mu_t-\sigma_t$ are assigned close-to-zero weights and thus considered dropped in the analysis. Here, $\mu_t / \sigma_t$ denotes the estimated \textit{mean}/\textit{std} of the overall confidence on unlabeled data, respectively.}
    \vspace{-0.2cm}
    \label{fig:dropped_top5}
\end{figure}

To address the trade-off between the quality and quantity of pseudo-labels in threshold-based pseudo-labeling, previous studies introduce dynamic threshold strategies that align with the evolving learning status of the model. For example, FlexMatch~\cite{zhang2021flexmatch} utilizes the number of confident pseudo-labels to estimate the learning difficulty for each class, subsequently mapping the predefined threshold to class-specific thresholds based on the determined difficulty levels. Additionally, FreeMatch~\cite{wang2022freematch} leverages the average confidence of unlabeled data to establish a dynamic global threshold. Besides, SoftMatch~\cite{chen2023softmatch} employs a Gaussian function representing the disparity between sample-specific and global confidence to model sample weights, assigning a positive weight to each unlabeled sample. However, samples with confidence significantly lower than the global confidence receive close-to-zero weights, essentially treated as if they were discarded. This characteristic positions SoftMatch as a variant of the threshold-based method. While the aforementioned algorithms effectively employ pseudo-labels to assess the learning status, the impact of biased data sampling or potential inter-class similarities can significantly influence the predictions of unlabeled data. Consequently, a pivotal question arises: \textit{Can we incorporate additional evidence along with pseudo-labels to achieve a more accurate estimation of the learning status?}

While an improved threshold scheme can enhance the utilization of unlabeled data, a portion of unlabeled samples still face exclusion. This raises another question: \textit{Can pseudo-labels assigned lower confidence provide valuable semantic guidance?} To address this concern, we examine the pseudo-label quality of previous algorithms. In the case of CIFAR-10~\cite{krizhevsky2009learning} with 40 labeled samples, more than half of the dropped pseudo-labels prove to be correct, as depicted in Figure~\ref{dropped_pseudo_acc}. Furthermore, Figure~\ref{top5_pseudo_acc} illustrates that the top-5 accuracy of the pseudo-labels reaches 100\% within just a few thousand iterations. Accordingly, pseudo-labels with lower confidence can eliminate false options~(\textit{e.g.}, bottom-5 classes) and provide effective supervision signals.

Motivated by the aforementioned questions, this paper introduces AllMatch, a novel SSL model designed to enhance learning status estimation and provide semantic guidance for all unlabeled data. Specifically, AllMatch proposes a class-specific adaptive threshold~(CAT) strategy, comprising global estimation and local adjustment steps, to achieve an improved characterization of the model's learning status. The global estimation step, similar to FreeMatch, employs the average confidence of unlabeled data as the global threshold. The ensuing local adjustment step utilizes the classifier weights to estimate the learning status of each class, adaptively decreasing thresholds for classes facing challenges. As illustrated in Figure~\ref{fig:indicators}(b, c) and Figure~\ref{fig:indicators}(f, g), CAT outperforms previous approaches in terms of the utilization ratio and pseudo-label accuracy of unlabeled samples. Besides, in response to the underutilization of unlabeled data resulting from the exclusion of low-confidence pseudo-labels, AllMatch introduces a binary classification consistency (BCC) regulation strategy to exploit the latent potential within such pseudo-labels. In essence, the BCC regulation divides the class space into candidate and negative classes, encouraging consistent candidate-negative division across diverse perturbed views of the same sample to eliminate negative options. The candidate class for each sample corresponds to its top-k predictions, considering the impressive top-k performance of various algorithms. Note that the parameter $k$ is dynamically determined based on varying sample-specific learning status and evolving model performance. As depicted in Figure~\ref{fig:indicators}(c, d) and Figure~\ref{fig:indicators}(g, h), the BCC regulation effectively identifies candidate classes for unlabeled samples and achieves a 100\% utilization ratio for the unlabeled data. Overall, our contributions can be summarized as follows:

(1) We revisit existing SSL algorithms and raise two questions: how to develop an effective threshold mechanism and how to utilize the low-confidence pseudo-labels.

(2) We propose the class-specific adaptive threshold mechanism, which employs pseudo-labels and classifier weights to estimate global and class-specific learning status respectively.

(3) We design the binary classification consistency regulation to provide supervision signals for all unlabeled samples.

(4) We conduct experiments on multiple benchmarks, considering both balanced and imbalanced settings. The results indicate that AllMatch achieves state-of-the-art performance.

\section{Related Work}
Consistency regulation and pseudo labelling~\cite{laine2016temporal,sajjadi2016regularization,tarvainen2017mean} are fundamental approaches in SSL. The former encourages consistent predictions across different perturbed views of the same sample while the latter assigns pseudo-labels to unlabeled samples. Among established techniques, FixMatch~\cite{sohn2020fixmatch} perfectly combines both techniques, establishing an effective SSL paradigm. Specifically, FixMatch uses the predictions of weakly augmented samples as pseudo-labels and minimize their divergence from the predictions of the corresponding strongly augmented views.

To ensure high-quality pseudo-labels, FixMatch utilizes a high constant threshold throughout training to filter out potentially incorrect pseudo-labels. However, this strategy results in the underutilization of unlabeled data. To address this issue, FlexMatch~\cite{zhang2021flexmatch} draws inspiration from curriculum learning~\cite{bengio2009curriculum}, mapping the predefined threshold to class-specific thresholds based on the learning status of each category. Dash~\cite{xu2021dash} defines the threshold based on the loss of labeled data, eliminating the empirical threshold parameter. FreeMatch~\cite{wang2022freematch} employs the average confidence on unlabeled data as the adaptive global threshold. SoftMatch~\cite{chen2023softmatch} estimates sample weights by a dynamic Gaussian function, maintaining soft margins between unlabeled samples of different confidence levels. In addition to the threshold-based algorithms, CoMatch~\cite{li2021comatch} and SimMatch~\cite{zheng2022simmatch} leverage contrastive loss to impose sample-level constraints on all unlabeled data. In contrast, AllMatch combines the advantages of both threshold-based and contrastive-based methods by introducing CAT, a learning-status-aware threshold strategy, and BCC regulation, the semantic-level supervision for the entire unlabeled set.

Concurrent with our work, FullMatch~\cite{chen2023boosting} also introduces semantic guidance for all unlabeled samples. Specifically, FullMatch compares the predictions of weakly and strongly augmented samples to identify negative classes. In contrast, AllMatch identifies candidate classes by comparing sample and global top-k confidence, thus considering the learning state of both individual sample and model in this process. Additionally, FullMatch assigns low probability~(similar to label smoothing) for the negative classes in the optimization objective, while AllMatch directly encourages consistent candidate-negative division for all unlabeled samples, thus exhibiting better consistency with the unsupervised loss.

In addition to consistency regulation and pseudo-labeling, entropy-based regulation is another widely adopted strategy. Entropy minimization~\cite{grandvalet2004semi} promotes high-confidence predictions during training. Maximizing the entropy of the expectation over all samples~\cite{krause2010discriminative,arazo2020pseudo,zhao2022dc} introduces the concept of \textit{fairness}, which encourages the model to predict each class with equal frequency. Specifically, distribution alignment~(DA)~\cite{berthelot2019remixmatch} and uniform alignment~(UA)~\cite{chen2023softmatch} are prevailing strategies for achieving fairness in SSL, which adjusts the pseudo-label based on the overall predictions on the unlabeled data.

\begin{figure*}[t]
    \centering
    \includegraphics[width=\textwidth]{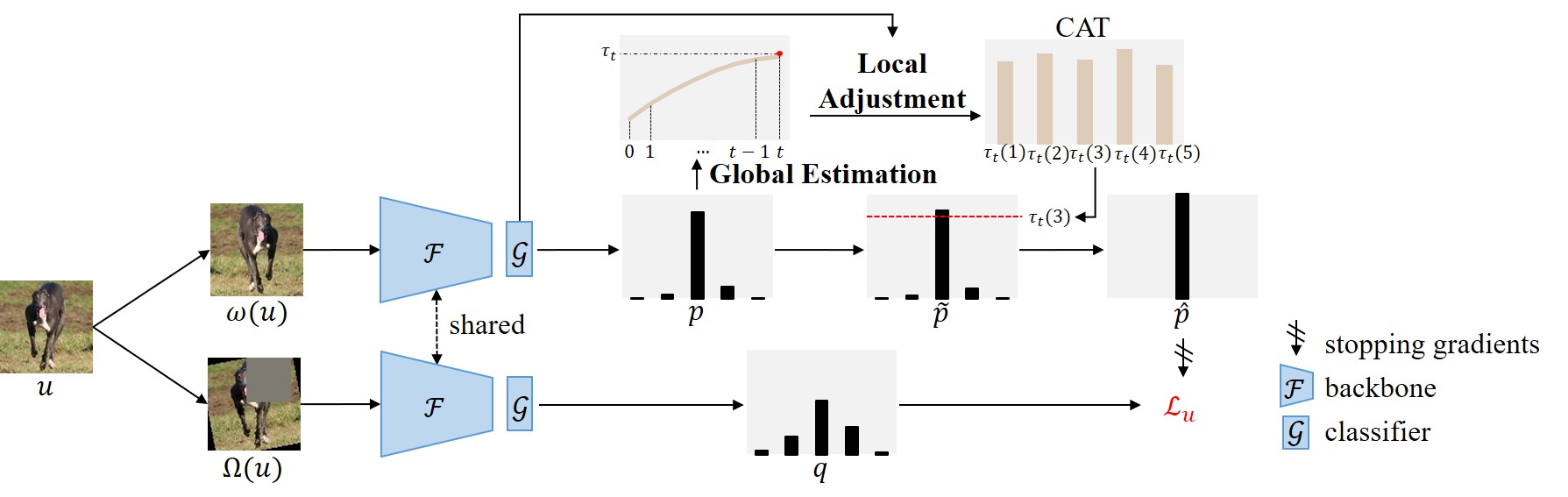}
    \caption{Pipeline of the class-specific adaptive threshold~(CAT) mechanism. CAT employs the average confidence on unlabeled data as the global threshold and subsequently utilizes the classifier weights to establish class-specific thresholds.}
    \vspace{-0.2cm}
    \label{fig:CAT}
\end{figure*}

\section{Methodology}
\subsection{Preliminary}
We begin by reviewing the widely adopted SSL framework. Let $\mathcal{D}_L=\{(x_i, y_i)\}_{i=1}^{N_L}$ and $\mathcal{D}_U=\{u_i\}_{i=1}^{N_U}$ represent the labeled and unlabeled datasets, respectively. Here, $x_i$ and $u_i$ denote the labeled and unlabeled training samples, and $y_i$ represents the one-hot label for the labeled sample $x_i$. We denote the prediction of sample $x$ as $p(y|x)$. Given a batch of labeled and unlabeled data, the model is optimized with the objective $\mathcal{L} = \mathcal{L}_s + \lambda_u \mathcal{L}_u$. Here, $\mathcal{L}_s$ represents the cross-entropy loss~($\mathcal{H}$) for the labeled batch of size $B_L$.
\begin{align}
    \mathcal{L}_s = \frac{1}{B_L}\sum_{i=1}^{B_L} \mathcal{H}(y_i, p(y|x_i))
\end{align}
$\mathcal{L}_u$ indicates the consistency regulation between the prediction of the strongly augmented view $\Omega(u)$ and the pseudo-label derived from the corresponding weakly augmented view $\omega(u)$. To filter out incorrect pseudo-labels, FixMatch~\cite{sohn2020fixmatch} introduces a predefined threshold $\tau$. Specifically, $\mathcal{L}_u$ is defined as follows.
\begin{align}
    \mathcal{L}_u &= \frac{1}{B_U}\sum_{i=1}^{B_U} \lambda(\tilde{p}_i) \mathcal{H}(\hat{p}_i, q_i) \\ 
    \lambda(p) &= 
    \begin{cases}
        1 & \textit{if }\textit{max}(p) \ge \tau \\
        0 & \textit{otherwise}
    \end{cases}
\end{align}
Here, $\tilde{p}_i$ is the abbreviation for $\textit{DA}(p(y|\omega(u_i)))$, where \textit{DA} indicates the distribution alignment strategy~\cite{berthelot2019remixmatch}. $\hat{p}_i$ represents the one-hot pseudo-label obtained from \textit{argmax}$(\tilde{p}_i)$. Moreover, $q_i$ is the abbreviation for $p(y|\Omega(u_i))$. Lastly, $B_U$ corresponds to the batch size of unlabeled data. 

\begin{figure*}[t]
    \centering
    \includegraphics[width=\textwidth]{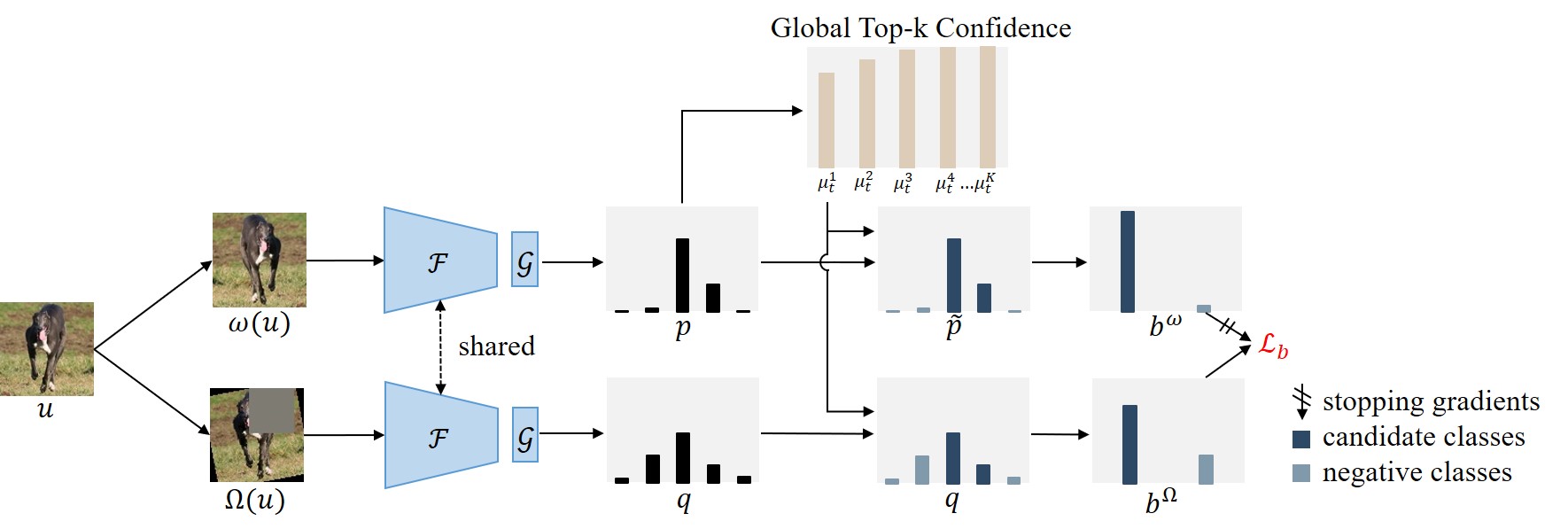}
    \caption{Pipeline of the binary classification consistency~(BCC) regulation. The module compares global and local top-k confidence to identify the candidate and negative classes for each unlabeled sample. Moreover, it encourages consistent candidate-negative division between different perturbed views of the same sample, thereby introducing supervision signals for all unlabeled samples.}
    \label{fig:BCC}
\end{figure*}

\subsection{Class-Specific Adapative Threshold}
Previous studies~\cite{zhang2021flexmatch,wang2022freematch} have demonstrated that the threshold should be aligned with the evolving learning status of the model. To achieve this, these approaches leverage predictions on unlabeled data to establish the dynamic threshold. In this paper, we unveil the ability of the classifier weights to differentiate the learning status of each class. By combining pseudo-labels and classifier weights, we introduce a class-specific adaptive threshold (CAT) mechanism. As depicted in Figure~\ref{fig:CAT}, CAT comprises the global estimation and local adjustment steps. The following parts provide a detailed description of these two steps.

\textbf{Global Estimation.} The global estimation step learns from FreeMatch~\cite{wang2022freematch} and evaluates the overall learning status of the model. Given that deep neural networks tend to prioritize fitting easier samples before memorizing harder and noisier ones, a lower threshold is necessary during early training stages to incorporate more correct pseudo-labels. Conversely, as training progresses, a higher threshold is required to filter out incorrect pseudo-labels. Given that the cross-entropy loss encourages confident predictions, the average confidence of the unlabeled set captures information from all unlabeled data and steadily increases throughout training, thereby reflecting the overall learning status. However, making predictions for the entire unlabeled set at each time step incurs significant computational costs. Accordingly, we employ the mean confidence of the current batch as an estimation and update it using exponential moving average~(EMA). Specifically, the global learning status estimation at $t$-th iteration, denoted as $\tau_t$, can be computed as follows:
\begin{align}
    \tau_t &=
    \begin{cases}
        \frac{1}{C} & \textit{if } t=0 \\
        m\tau_{t-1} + (1-m)\frac{1}{B_U}\sum\limits_{i=1}^{B_U}\textit{max}(p_i) & \textit{otherwise}
    \end{cases}
\end{align}
Here, $p_i$ represents $p(y|\omega(u_i))$, $m$ denotes the momentum decay, and $C$ corresponds to the number of classes. 

\textbf{Local Adjustment.} Due to the inherent variations in learning difficulty among different classes and the stochastic nature of parameter initialization, the model's learning status varies across categories. To address this issue, we introduce the local adjustment step, which makes the model pay more attention to the underfitting classes by decreasing their thresholds. Specifically, our study reveals that the L2 norm of the classifier weights provides insights into class-specific learning status. The reasons are explained as follows.

Firstly, let $\mathcal{M}=\mathcal{G}\circ \mathcal{F}$ denote the model, where $\mathcal{F}$ and $\mathcal{G}$ correspond to the encoder and the one-layer classifier, respectively. Given the feature vector $f=\mathcal{F}(u) \in R^d$ of the unlabeled sample $u$, the predicted logits $z\in R^C$ can be computed as $z=\mathcal{G}(f)=fW^T$, where $W \in R^{C \times d}$ denotes the weight matrix of the classifier. The bias term is dropped as it typically has a negligible impact on the results. Consequently, the logits of class $c$ can be expressed as $z_c = ||f||\cdot||W_c||\cdot\textit{cos}(\theta)$, where $||\cdot||$ denotes the L2 norm, and $W_c$ corresponds to the weight of class $c$. Accordingly, the model tends to produce logits with larger absolute values for categories with larger weight norms, implying that classes with larger weight norms exhibit a preferred learning status. 

Secondly, \cite{kang2019decoupling} reveals a positive correlation between the weight norm $||W_c||$ and $n_c$, the number of samples in class $c$. Given the abundant unlabeled data and limited labeled data, $n_c$ can be approximated as the number of unlabeled samples classified into class $c$ with confidence scores surpassing the threshold. Hence, a larger weight norm suggests that more samples are confidently categorized into the class, signifying a better learning status.

According to the above analysis, the L2 norm of classifier weights characterizes the learning status of each class. Consequently, the local adjustment step leverages this indicator to establish the mapping from the global threshold to class-specific thresholds. Specifically, we linearly scale the threshold for each class based on the deviation of its learning status from the optimal learning status. As such, the threshold for class $c$ at $t$-th iteration, denoted as $\tau_t(c)$, can be computed as follows.
\begin{align}
    \tau_t(c) = \tau_t \cdot \frac{||W_c||}{\textit{max}\{||W_c||: c \in [1, \cdots, C]\}}
\end{align}
Moreover, to ensure a stable estimation of the learning status, we employ the classifier weights obtained from the EMA model. Notably, in contrast to FlexMatch, which maintains an additional list for recording the selected pseudo-label of each sample, the proposed CAT refrains from storing any sample-specific information during training. This eliminates the indexing budget concerns on large-scale datasets. With CAT incorporated, the mask for unlabeled samples in $\mathcal{L}_u$ can be expressed as follows.
\begin{align}
    \lambda(p) &= 
    \begin{cases}
        1 & \textit{if }\textit{max}(p) \ge \tau_t(\textit{argmax}(p)) \\
        0 & \textit{otherwise}
    \end{cases}
\end{align}

\subsection{Binary Classification Consistency Regulation}
While the proposed class-specific adaptive threshold alleviates the underutilization of unlabeled data, a substantial number of pseudo-labels continue to be discarded. As illustrated in Figure~\ref{fig:dropped_top5}, in the case of CIFAR-10 with 40 labeled samples, the top-5 accuracy of pseudo-labels effortlessly achieves 100\% regardless of the adopted algorithm. In other words, pseudo-labels assigned lower confidence contribute to identifying candidate classes~(\textit{e.g.,} top-k predictions) and excluding negative options~(\textit{e.g.,} classes not included in top-k predictions). Motivated by these observations and the consistency regulation technique, we propose the binary classification consistency (BCC) regulation, whose overview is shown in Figure~\ref{fig:BCC}. In a nutshell, the strategy introduces semantic supervision for all unlabeled data by encouraging consistent candidate-negative division across diverse perturbed views of the same sample. The details are described as follows. 

Given the impressive top-k pseudo-label accuracy obtained by numerous algorithms, the BCC regulation adopts the top-k predictions of each unlabeled sample as its candidate classes and the rest as the negative options. Thus, the candidate-negative division is simplified to the selection of the parameter $k$. Moreover, considering the variations in learning difficulty among different samples and the evolving performance of the model, the candidate-negative division for each sample should be determined based on both individual and global learning status. To achieve this, the BCC regulation first computes sample-specific top-k confidence and the global top-k confidence of the entire unlabeled set. Specifically, let $p_i^k$ denote the top-k probability of sample $u_i$, and $\mu_t^k$ represent the global top-k probability at $t$-th iteration. The global top-k confidence can be estimated by the exponential moving average~(EMA) of the average top-k confidence at each time step.
\begin{align}
    p_i^k &= \sum_{j=1}^k p_{i, c_j} \quad (p_{i, c_1} \ge p_{i, c_2} \ge \cdots) \\
    \mu_t^k &=
    \begin{cases}
        \frac{k}{C} & \textit{if } t=0 \\
        m\mu_{t-1}^k + (1-m)\frac{1}{B_U}\sum_{i=1}^{B_U} p_i^k & \textit{otherwise}
    \end{cases}
\end{align}
Here, $c_1,\dots,c_k$ represent the $k$ classes assigned the highest probability in $p_i$. With the global top-k confidence determined, the number of candidate classes for each unlabeled sample is defined as the minimum value that makes individual top-k confidence higher than global top-k confidence. Particularly, the candidate class for confident unlabeled samples is defined as the pseudo-label. Accordingly, the number of candidate classes $k_i$ for sample $u_i$ can be expressed as follows.
\begin{align}
    k_i &= 
    \begin{cases}
        1  & \textit{if } \lambda(\tilde{p}_i) = 1 \\
        \textit{min}(\textit{min}\{k: \tilde{p}_i^k \ge \mu_t^k\}, K) & \textit{otherwise}
    \end{cases}
\end{align}
where $K$ is the upper bound for the number of candidate classes to prevent trivial candidate-negative division. With the division obtained, the candidate and negative probabilities for the weakly~($b_i^\omega$) and strongly~($b_i^\Omega$) perturbed views of the unlabeled sample $u_i$ can be calculated as follows.
\begin{align}
    b_i^\omega &= [\sum_{j=1}^{k_i} \tilde{p}_{i, c_j}, \sum_{j=k_i+1}^C \tilde{p}_{i, c_j}] \quad (\tilde{p}_{i, c_1} \ge \tilde{p}_{i, c_2} \ge \cdots) \\
    b_i^\Omega &= [\sum_{j=1}^{k_i} q_{i, c_j}, \sum_{j=k_i+1}^C q_{i, c_j}]
\end{align}
Here, $c_1,\dots,c_{k_i}$ represents the $k_i$ classes assigned the highest probability in $\tilde{p}_i$. Finally, the BCC regulation for a batch of unlabeled data can be calculated as follows:
\begin{align}
    \mathcal{L}_b = \frac{1}{B_U}\sum_{i=1}^{B_U}\mathcal{H}(b_i^\omega, b_i^\Omega)
\end{align}

\subsection{Overall Objective}
The overall objective of AllMatch is defined as the weighted sum of all semantic-level supervision.
\begin{align}
    \mathcal{L} = \mathcal{L}_s + \lambda_u \mathcal{L}_u + \lambda_b \mathcal{L}_b
\end{align}
$\lambda_u$ and $\lambda_b$ denote the weights to balance different supervision signals. For all experiments, we set both $\lambda_u$ and $\lambda_b$ to 1.0.

\begin{table*}[t]
    \scriptsize
    \centering
    \tabcolsep=0.07cm
    \begin{tabular}{c|cccc|ccc|cc|cc}
    \toprule
    Datasets & \multicolumn{4}{c}{CIFAR-10} \vline & \multicolumn{3}{c}{CIFAR-100} \vline & \multicolumn{2}{c}{SVHN} \vline & \multicolumn{2}{c}{STL-10} \\
    \midrule
    \# Label & 10 & 40 & 250 & 4000 & 400 & 2500 & 10000 & 40 & 1000 & 40 & 1000\\
    \midrule
    MixMatch~\cite{berthelot2019mixmatch} & 34.24{\tiny±7.06} & 63.81{\tiny ±6.48} & 86.37{\tiny ±0.59} & 93.34{\tiny ±0.26} & 32.41{\tiny ±0.66} & 60.24{\tiny ±0.48} & 72.22{\tiny ±0.29} & 69.40{\tiny ±8.39} & 96.31{\tiny ±0.37} & 45.07{\tiny ±0.96} & 78.30{\tiny ±0.68} \\
    ReMixMatch~\cite{berthelot2019remixmatch} & 79.23{\tiny±7.48} & 90.12{\tiny ±1.03} & 93.70{\tiny ±0.05} & 95.16{\tiny ±0.01} & 57.25{\tiny ±1.05} & \underline{73.97{\tiny ±0.35}} & \textbf{79.98{\tiny ±0.27}} & 75.96{\tiny ±9.13} & 94.84{\tiny ±0.31} & 67.88{\tiny ±6.24} & 93.26{\tiny ±0.14} \\
    UDA~\cite{xie2020unsupervised} & 65.47{\tiny±10.69} & 89.38{\tiny ±3.75} & 94.84{\tiny ±0.06} & 95.71{\tiny ±0.07} & 53.61{\tiny ±1.59} & 72.27{\tiny ±0.21} & 77.51{\tiny ±0.23} & 94.88{\tiny ±4.27} & \underline{98.11{\tiny ±0.01}} & 62.58{\tiny ±8.44} & 93.36{\tiny ±0.17} \\
    FixMatch~\cite{sohn2020fixmatch} & 82.09{\tiny±2.48} & 92.53{\tiny ±0.28} & 95.14{\tiny ±0.05} & 95.79{\tiny ±0.08} & 53.58{\tiny ±0.82} & 71.97{\tiny ±0.16} & 77.80{\tiny ±0.12} & 96.19{\tiny ±1.18} & 98.04{\tiny ±0.03} & 64.03{\tiny ±4.14} & 93.75{\tiny ±0.33} \\
    Dash~\cite{xu2021dash} & 73.72{\tiny±14.09} & 91.07{\tiny ±3.11} & 94.84{\tiny ±0.23} & 95.64{\tiny ±0.11} & 55.18{\tiny ±0.96} & 72.85{\tiny ±0.22} & 78.12{\tiny ±0.07} & \underline{97.81{\tiny ±0.18}} & 98.03{\tiny ±0.01} & 65.48{\tiny ±4.30} & 93.61{\tiny ±0.56} \\
    MPL~\cite{pham2021meta} & 76.45{\tiny±6.01} & 93.38{\tiny ±0.91} & 94.24{\tiny ±0.24} & 95.45{\tiny ±0.04} & 53.74{\tiny ±1.84} & 72.29{\tiny ±0.19} & 78.26{\tiny ±0.09} & 90.67{\tiny ±8.02} & 97.72{\tiny ±0.02} & 64.24{\tiny ±4.83} & 93.34{\tiny ±0.08} \\
    FullMatch~\cite{chen2023boosting} & - & 94.11{\tiny ±1.01} & \textbf{95.36{\tiny ±0.12}} & \textbf{96.25{\tiny ±0.08}} & 59.42{\tiny ±1.40} & 73.06{\tiny ±0.40} & 78.56{\tiny ±0.10} & 97.65{\tiny ±0.10} & 98.01{\tiny ±0.03} & - & 94.26{\tiny ±0.09}   \\
    FlexMatch~\cite{zhang2021flexmatch} & 90.02{\tiny±1.95} & 95.03{\tiny ±0.06} & 95.02{\tiny ±0.09} & 95.81{\tiny ±0.01} & 60.06{\tiny ±1.62} & 73.51{\tiny ±0.20} & 78.10{\tiny ±0.15} & 91.81{\tiny ±3.20} & 93.28{\tiny ±0.30} & 76.65{\tiny ±2.07} & 94.23{\tiny ±0.18} \\
    SoftMatch~\cite{chen2023softmatch} & \underline{93.04\tiny±1.27} & 95.09{\tiny ±0.12} & 95.18{\tiny ±0.09} & 95.96{\tiny ±0.02} & \underline{62.90{\tiny ±0.77}} & 73.34{\tiny ±0.25} & 77.97{\tiny ±0.03} & 97.67{\tiny ±0.25} & 97.99{\tiny ±0.01} & \underline{85.28{\tiny ±1.93}} & 94.27{\tiny ±0.24} \\
    FreeMatch~\cite{wang2022freematch} & 92.09{\tiny±0.85} & \underline{95.10{\tiny ±0.04}} & 95.12{\tiny ±0.18} & 95.90{\tiny ±0.02} & 62.02{\tiny ±0.42} & 73.53{\tiny ±0.20} & 78.32{\tiny ±0.03} & \textbf{98.03{\tiny ±0.02}} & 98.04{\tiny ±0.03} & 84.68{\tiny ±0.82} & \underline{94.37{\tiny ±0.15}} \\
    \midrule
    AllMatch & \textbf{94.91{\tiny±0.27}} & \textbf{95.20{\tiny ±0.08}}& \underline{95.28{\tiny ±0.06}} & \underline{96.14{\tiny ±0.04}} & \textbf{63.56{\tiny±0.78}} & \textbf{74.16{\tiny ±0.12}} & \underline{78.66{\tiny ±0.09}} & 97.56{\tiny±0.12} & \textbf{98.15{\tiny ±0.06}} & \textbf{88.14{\tiny±1.22}} & \textbf{94.91{\tiny ±0.26}} \\
    \bottomrule
    \end{tabular}
    \caption{Top-1 accuracy~(\%) on CIFAR-10, CIFAR-100, SVHN, and STL-10 with varying number of labeled samples. \textbf{Bold} indicates the best performance, and \underline{underline} denotes the second best performance.}
    \label{tab:balanced}
\end{table*}

\begin{table}[t]
    \scriptsize
    \centering
    \tabcolsep=0.1cm
    \begin{tabular}{c|c c}
    \toprule
    Datasets & \multicolumn{2}{c}{ImageNet}\\
    \midrule
    \# Label & 100k & 400k\\
    \midrule
    FixMatch~\cite{sohn2020fixmatch} & 56.34 & 67.72 \\
    FlexMatch~\cite{zhang2021flexmatch} & 58.15 & 68.69 \\
    SoftMatch~\cite{chen2023softmatch} & \underline{59.48} & \underline{70.51} \\
    FreeMatch~\cite{wang2022freematch} & 59.43 & 69.48 \\
    \midrule
    AllMatch & \textbf{59.99} & \textbf{70.82} \\
    \bottomrule
    \end{tabular}
    \caption{Top-1 accuracy~(\%) on ImageNet.}
    \label{tab:imagenet}
\end{table}

\section{Experiments}

\subsection{Balanced Semi-Supervised Learning}
\textbf{Settings.} For balanced image classification, we conduct experiments on CIFAR-10/100~\cite{krizhevsky2009learning}, SVHN~\cite{netzer2011reading}, STL-10~\cite{coates2011analysis}, and ImageNet~\cite{deng2009imagenet} with various numbers of labeled data, where the class distribution of the labeled data is balanced. To ensure fair comparisons, we employ the unified codebase TorchSSL~\cite{zhang2021flexmatch} to evaluate all methods. Regarding the backbone architecture, we follow previous studies and use specific models for different datasets: WRN-28-2~\cite{zagoruyko2016wide} for CIFAR-10 and SVHN, WRN-28-8 for CIFAR-100, WRN-37-2~\cite{zhou2020time} for STL-10, and ResNet-50~\cite{he2016deep} for ImageNet. The batch sizes $B_L$ and $B_U$ are set to 128 and 128 for ImageNet and 64 and 448 for the remaining datasets. AllMatch is trained using the SGD optimizer with an initial learning rate of 0.03 and a momentum decay of 0.9. The learning rate is adjusted by a cosine decay scheduler over a total of $2^{20}$ iterations. We set $m$ to 0.999 and generate the EMA model with a momentum decay of 0.999 for inference. The upper bound $K$ is set to 20 for ImageNet and 10 for the other datasets. For SVHN, CIFAR-10 with 10 labels, and STL-10 with 40 labels, we constrain the threshold within the range of [0.9, 1.0] to prevent overfitting noisy pseudo-labels in the early training stages. To account for randomness, we repeat each experiment three times and report the mean and standard deviation of the top-1 accuracy. \textit{Detailed implementation and data processing are listed in Appendix C}.

\textbf{Performance.} Table~\ref{tab:balanced} presents the top-1 accuracy on CIFAR-10/100, SVHN, and STL-10 with various numbers of labeled samples. The performance on ImageNet is reported in Table~\ref{tab:imagenet}. The experimental results demonstrate that AllMatch achieves state-of-the-art performance on almost all datasets. For CIFAR-10, AllMatch outperforms FullMatch with only 40 available labels, and performs comparably to FullMatch when there are 250 or 4000 labels. Moreover, regarding CIFAR-100, AllMatch outperforms ReMixMatch when only 400 or 2500 labels are available, while the latter achieves better performance when 10000 labels are available. The competitive results obtained by ReMixMatch mainly stem from the Mixup technique~\cite{zhang2017mixup} and the additional self-supervised learning part. Furthermore, AllMatch exhibits substantial advantages over previous algorithms when dealing with extremely limited labeled samples. Specifically, the approach surpasses the second-best counterpart by 1.87\% on CIFAR-10 with 10 labels, 0.66\% on CIFAR-100 with 400 labels, and 2.86\% on STL-10 with 40 labels. Particularly, STL-10 poses significant challenges due to its large unlabeled set that comprises 100k images. Accordingly, the impressive improvement obtained on STL-10 highlights the potential of AllMatch to be deployed in real-world applications.

\begin{table}[t]
    \scriptsize
    \centering
    \tabcolsep=0.05cm
    \begin{tabular}{c|ccc|ccc}
    \toprule
    Dataset & \multicolumn{3}{c}{CIFAR-10-LT} \vline & \multicolumn{3}{c}{CIFAR-100-LT} \\
    \midrule
    $\gamma$ & 50 & 100 & 150 & 20 & 50 & 100 \\
    \midrule
    FixMatch & 81.54{\tiny ±0.30} & 74.89{\tiny ±1.20} & 70.38{\tiny ±0.88} & 49.58{\tiny ±0.78} & 42.11{\tiny ±0.33} & 37.60{\tiny ±0.48} \\
    FlexMatch & 81.87{\tiny ±0.19} & 74.49{\tiny ±0.92} & 70.20{\tiny ±0.36} & 50.89{\tiny ±0.60} & 42.80{\tiny ±0.39} & 37.30{\tiny ±0.47} \\
    SoftMatch & \underline{83.45{\tiny ±0.29}} & \underline{77.07{\tiny ±0.37}} & \underline{72.60{\tiny ±0.46}} & \underline{51.91{\tiny ±0.55}} & \underline{43.76{\tiny ±0.51}} & \underline{38.92{\tiny ±0.81}} \\
    FreeMatch & 83.30{\tiny ±0.33} & 75.96{\tiny ±0.49} & 71.20{\tiny ±0.64} & 51.60{\tiny ±0.91} & 42.95{\tiny ±0.48} & 37.50{\tiny ±0.23}\\
    \midrule
    AllMatch & \textbf{84.21{\tiny ±0.33}} & \textbf{78.76{\tiny ±0.33}} & \textbf{74.25{\tiny ±0.37}}& \textbf{52.52{\tiny ±0.33}}& \textbf{44.10{\tiny ±0.36}} & \textbf{39.09{\tiny ±0.27}}\\
    \bottomrule
    \end{tabular}
    \caption{Performance~(\%) on CIFAR-10-LT and CIFAR-100-LT.}
    \label{tab:imbalance}
\end{table}

\subsection{Imbalanced Semi-Supervised Learning}
\textbf{Settings.} We evaluate AllMatch in the context of imbalanced SSL, where both labeled and unlabeled data exhibit a long-tailed distribution. All experiments are conducted on the TorchSSL codebase. Following prior studies~\cite{lee2021abc,oh2022daso,wei2021crest,lai2022smoothed,fan2022cossl,chen2023softmatch}, we generate the labeled and unlabeled sets using the configurations of $N_c=N_1\cdot\gamma^{-\frac{c-1}{C-1}}$ and $M_c=M_1\cdot\gamma^{-\frac{c-1}{C-1}}$. Specifically, for CIFAR-10-LT, we set $N_1$ to $1500$, $M_1$ to $3000$, and $\gamma$ to range from 50 to 150. For CIFAR-100-LT, we set $N_1$ to $150$, $M_1$ to $300$, and $\gamma$ to range from 20 to 100. In all experiments, we employ WRN-28-2 as the backbone and utilize the Adam optimizer~\cite{kingma2014adam} with the weight decay of 4e-5. The batch sizes $B_L$ and $B_U$ are set to 64 and 128, respectively. The learning rate is initially set to 2e-3 and adjusted by a cosine decay scheduler during training. We repeat each experiment three times and report the overall performance. \textit{Detailed implementation is listed in Appendix C}.

\textbf{Performance.} In the context of imbalanced SSL, we compare AllMatch with several strong baselines, including FixMatch, FlexMatch, SoftMatch, and FreeMatch. The results in Table \ref{tab:imbalance} demonstrate that AllMatch achieves state-of-the-art performance on all benchmarks. It is particularly noteworthy that AllMatch outperforms the second-best approach by 1.69\% and 1.65\% at $\gamma$=100 and $\gamma$=150 on CIFAR-10-LT, respectively, highlighting its robustness in handling significant imbalances. Furthermore, as detailed in \textit{Appendix B}, AllMatch is compatible with existing imbalanced SSL algorithms, and their combination can further enhance resilience against severe imbalances. The consistently impressive performance observed in imbalanced SSL suggests that AllMatch can effectively address real-world challenges.

\begin{table}[t]
\scriptsize
    \centering
    \tabcolsep=0.1cm
    \begin{tabular}{ccc|cccc}
    \toprule
    GE & LA & BCC & CIFAR-10-10 & CIFAR-100-400 & STL-10-40 & CIFAR-10-LT-150\\
    \midrule
    & & & 82.09 & 53.58 & 64.03 & 70.38 \\
    \checkmark & & & 89.44 & 61.58 & 80.48 & 71.58 \\
    \checkmark & \checkmark & & 92.66 & 63.04 & 87.43 & 73.48 \\
    \checkmark & \checkmark & \checkmark & 94.91 & 63.56 & 88.14 & 74.25 \\
    \bottomrule
    \end{tabular}
    \caption{Ablation study~(\%). (GE: global estimation in CAT. LA: local adjustment in CAT. BCC: binary classification consistency.)}
    \label{tab:ablation}
\end{table}

\begin{table}[t]
    \scriptsize
    \centering
    \tabcolsep=0.08cm
    \begin{tabular}{cc|cccc}
    \toprule
    Threshold & BCC & CIFAR-10-10 & CIFAR-100-400 & STL-10-40 & CIFAR-10-LT-150\\
    \midrule
    FlexMatch & & 90.02 & 60.06 & 76.65 & 70.20 \\
    FreeMatch & & 92.09 & 62.02 & 84.68 & 71.20 \\
    SoftMatch & & \textbf{93.04} & \underline{62.90} & \underline{85.28} & \underline{72.60} \\
    AllMatch & & \underline{92.66} & \textbf{63.04} & \textbf{87.43} & \textbf{73.48} \\
    \midrule
    FlexMatch & \checkmark & 92.04 & 61.18 & 80.05 & 70.88 \\
    FreeMatch & \checkmark & 94.19 & 62.75 & 86.49 & 72.04 \\
    SoftMatch & \checkmark & \textbf{94.99} & \underline{63.21} & \underline{87.10} & \underline{73.20} \\
    AllMatch & \checkmark & \underline{94.91} & \textbf{63.56} & \textbf{88.14} & \textbf{74.25}\\
    \bottomrule
    \end{tabular}
    \caption{Threshold comparison study~(\%). SoftMatch assigns trivial weights to samples with confidence significantly lower than the global confidence, approximating itself as a threshold-based model.}
    \label{tab:threshold_ablation}
\end{table}

\subsection{Ablation Study}
In this part, we systematically evaluate each constituent component of AllMatch. \textit{Additionally, we provide the grid search of $K$~(upper bound for the number of candidate classes) and $\lambda_b$~(weight for the BCC regulation) in Appendix A.1 and A.2}.

\textbf{Component analysis.} We conduct an ablation study on four challenging datasets: CIFAR-10 with 10 labels, CIFAR-100 with 400 labels, STL-10 with 40 labels, and CIFAR-10-LT with an imbalance ratio of 150. For simplicity, we refer to the performance on the four benchmarks as $(a, b, c, d)$ in subsequent analysis. As shown in Table~\ref{tab:ablation}, the global estimation step in CAT~(line~2) promotes the performance by (7.35\%, 8.00\%, 16.45\%, 1.20\%) compared to the baseline model in line~1. The significant improvement highlights the crucial role of aligning the threshold with the model's global learning status. Furthermore, the local adjustment step in CAT~(line~3) yields additional gains of (3.22\%, 1.46\%, 6.95\%, 1.90\%), suggesting that it effectively captures class-specific learning difficulties and facilitates the learning of classes facing challenges. Additionally, the BCC regulation enables a 100\% utilization ratio of the unlabeled data and achieves the improvement of (2.25\%, 0.52\%, 0.71\%, 0.77\%). The substantial improvement observed on CIFAR-10 with 10 labels indicates the potential of the BCC regulation when dealing with extremely limited labeled data. Overall, the results in Table~\ref{tab:ablation} demonstrate the effectiveness of the proposed modules and the advantages of their combination in AllMatch.

\begin{figure*}[t]
    \centering
    \subfigure[Class-average threshold.]{
        \includegraphics[width=0.23\textwidth]{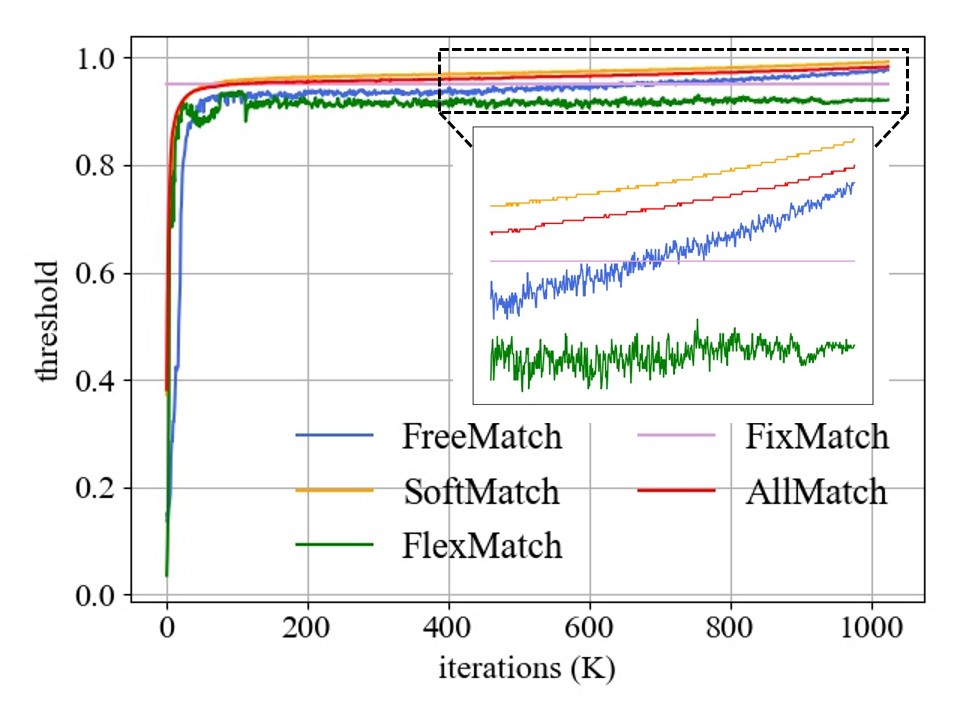}
        \label{threshold}
    }
    \subfigure[Selected pseudo-label \textit{acc}.]{
        \includegraphics[width=0.23\textwidth]{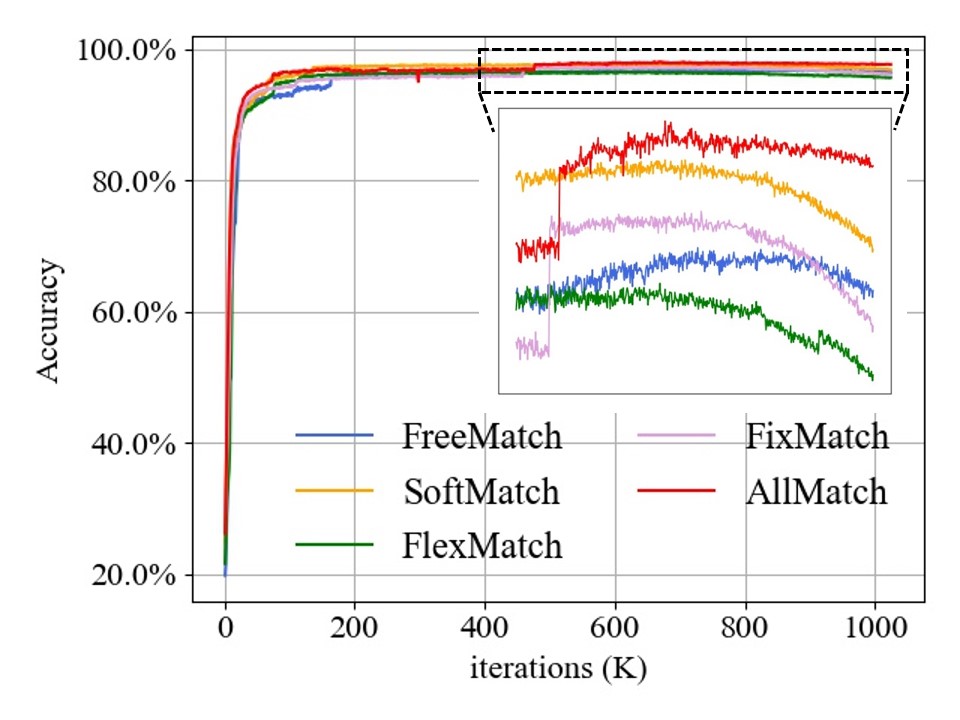}
        \label{select_pseudo_acc}
    }
    \subfigure[Utilization ratio.]{
        \includegraphics[width=0.23\textwidth]{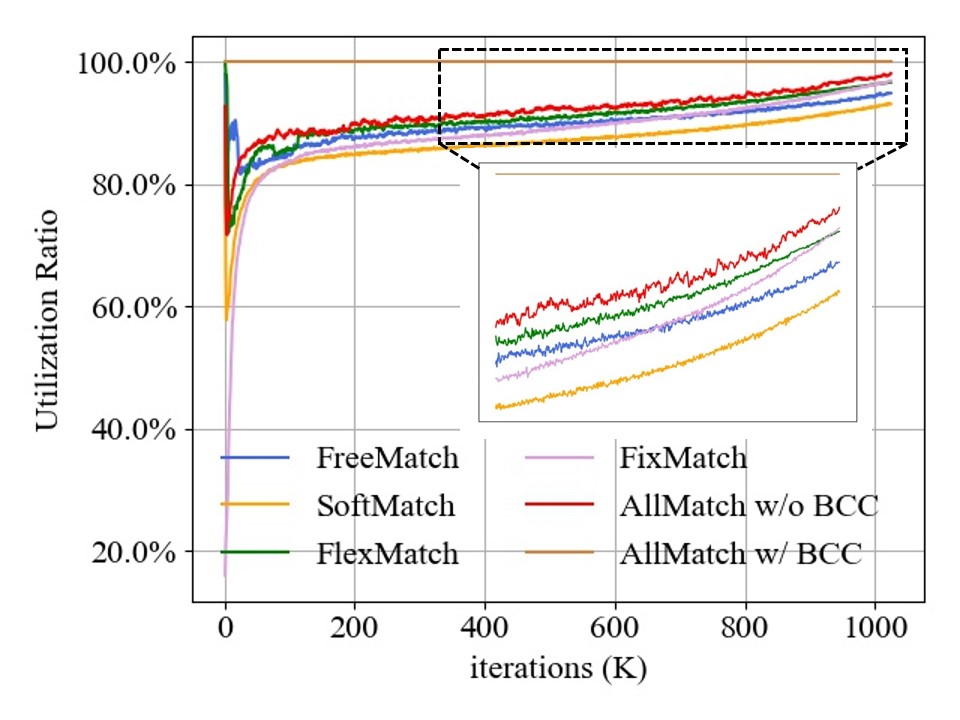}
        \label{utilization_ratio}
    }
    \subfigure[Binary pseudo-label \textit{acc}.]{
        \includegraphics[width=0.23\textwidth]{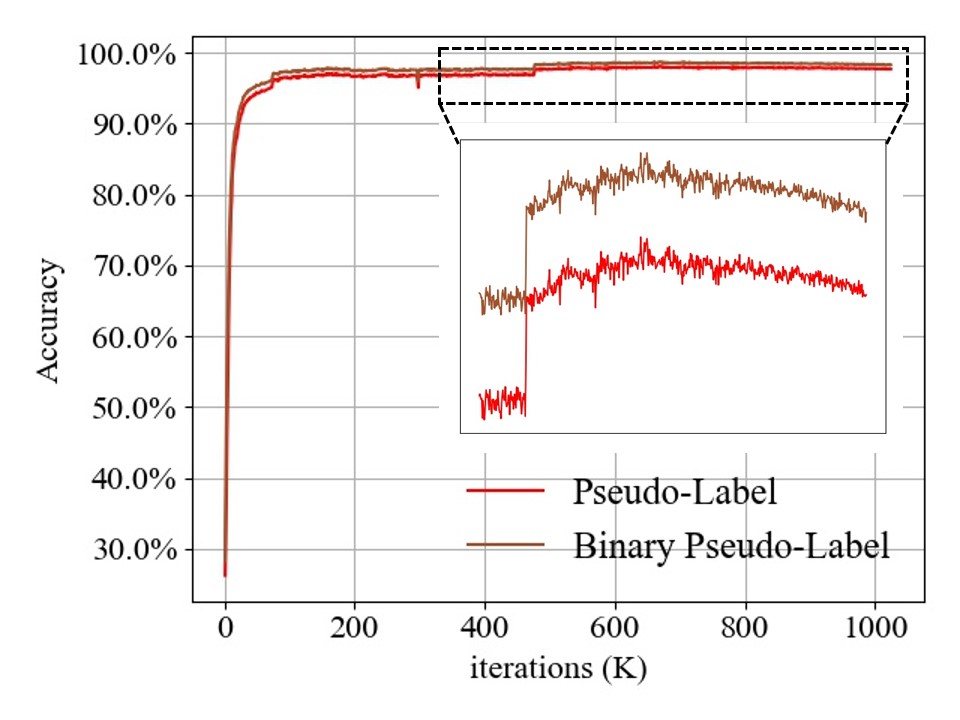}
        \label{binary_acc}
    }
    \subfigure[Class-average threshold.]{
        \includegraphics[width=0.23\textwidth]{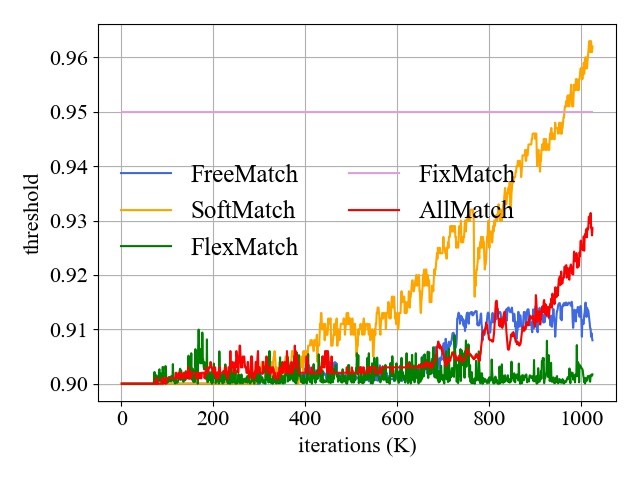}
        \label{threshold_stl10}
    }
    \subfigure[Selected pseudo-label \textit{acc}.]{
        \includegraphics[width=0.23\textwidth]{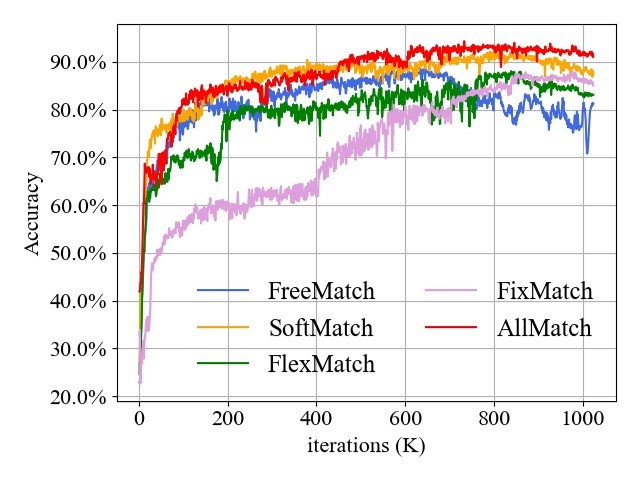}
        \label{select_pseudo_acc_stl10}
    }
    \subfigure[Utilization ratio.]{
        \includegraphics[width=0.23\textwidth]{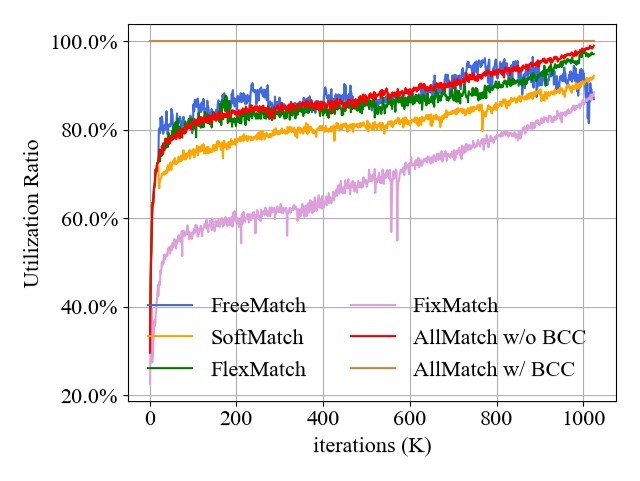}
        \label{utilization_ratio_stl10}
    }
    \subfigure[Binary pseudo-label \textit{acc}.]{
        \includegraphics[width=0.23\textwidth]{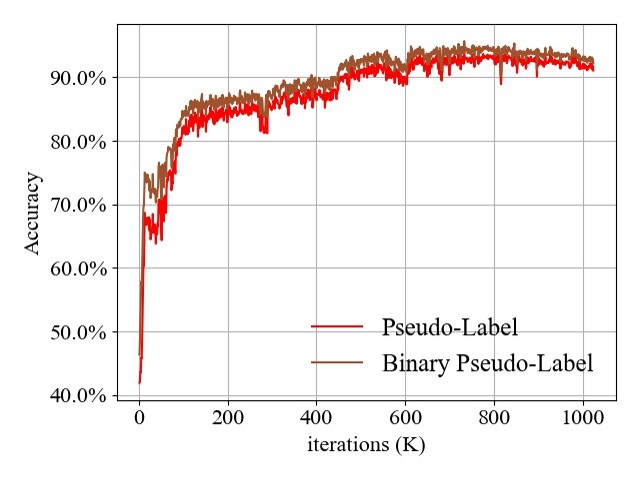}
        \label{binary_acc_stl10}
    }
    \caption{Learning process visualization on CIFAR-10-40~(a-d) and STL-10-40~(e-h). The threshold for STL-10-40 is restricted within [0.9, 1.0] to mitigate the adverse effects of noisy pseudo-labels. In the analysis of SoftMatch, we employ $\mu_t-\sigma_t$ as its threshold. Samples with a confidence lower than $\mu_t-\sigma_t$ are assigned negligible weights, essentially treated as if they were discarded. Consequently, the class-average threshold of SoftMatch is $\mu_t-\sigma_t$, and its selected pseudo-label \textit{acc} and utilization ratio can be computed like other threshold-based models. Here, $\mu_t / \sigma_t$ denotes the \textit{mean}/\textit{std} of the overall confidence on unlabeled data. \textit{Detailed analysis for SoftMatch is provided in Appendix A.3.}}
    \label{fig:indicators}
\end{figure*}

\textbf{Comparative study on threshold strategies.} We conduct a comparative analysis of existing threshold mechanisms in two aspects. Firstly, we directly compare the proposed CAT with the threshold strategies adopted in previous models. The results are presented in line~1-4 of Table~\ref{tab:threshold_ablation}. Secondly, we assess the threshold strategies within the AllMatch framework, \textit{i.e.,} combining existing threshold schemes with the BCC regulation, and the results are provided in line~5-8 of Table~\ref{tab:threshold_ablation}. From both perspectives, AllMatch outperforms previous models in most cases, indicating the effectiveness of the proposed CAT. Moreover, the BCC regulation further boosts the performance of prior methods, suggesting its impressive compatibility and contribution to eliminating false options.

\subsection{Quantitative Analysis}
To gain further insights into AllMatch, we present various training indicators on CIFAR-10 with 40 labels and STL-10 with 40 labels, as illustrated in Figure~\ref{fig:indicators}. \textit{Besides, the indicators on CIFAR-10 with 10 labels and CIFAR-100 with 400 labels are presented in Appendix A.3.} From Figure~\ref{threshold} and Figure~\ref{threshold_stl10}, it can be observed that the threshold exhibits the expected behavior, starting with a small value and gradually increasing thereafter. Moreover, AllMatch demonstrates a smoother threshold evolution in contrast to other class-specific threshold-based models, implying a preferred learning status estimation. Additionally, in comparison to previous algorithms, Figure~\ref{select_pseudo_acc} and Figure~\ref{utilization_ratio}, as well as Figure~\ref{select_pseudo_acc_stl10} and Figure~\ref{utilization_ratio_stl10}, indicate that AllMatch achieves improved pseudo-label accuracy and a higher utilization ratio for the unlabeled data. Notably, Figure~\ref{select_pseudo_acc} and Figure~\ref{select_pseudo_acc_stl10} demonstrate that previous models consistently suffer from overfitting noisy pseudo-labels in the later training stages, whereas AllMatch successfully mitigates this issue. Furthermore, as depicted in Figure~\ref{binary_acc} and Figure~\ref{binary_acc_stl10}, the accuracy of the candidate-negative division consistently outperforms pseudo-label accuracy, suggesting that the BCC regulation effectively identifies the candidate classes for all unlabeled data. Overall, the CAT precisely reflects the learning status of the model and the BCC regulation provides accurate supervision signals for all unlabeled samples. Additionally, to provide a comprehensive analysis of AllMatch, \textit{we present feature visualization~\cite{van2008visualizing} and confusion matrices on STL-10 with 40 labels in Appendix A.4 and A.5}.

\section{Conclusion}
This paper revisits prior SSL algorithms, focusing on two crucial questions: how to design an effective threshold mechanism and how to utilize the pseudo-labels assigned lower confidence. To address these challenges, we introduce two strategies named class-specific adaptive threshold~(CAT) and binary classification consistency~(BCC) regulation. CAT leverages predictions on unlabeled data and classifier weights to establish a threshold mechanism that aligns with the evolving learning status of each class. BCC regulation identifies candidate classes for each unlabeled sample and encourages consistent candidate-negative divisions across diverse perturbed views of the same sample. With these two modules incorporated, the proposed AllMatch maximizes the utilization of unlabeled data and achieves impressive pseudo-label accuracy. We conduct extensive experiments on multiple benchmarks, including both balanced and imbalanced settings. The results demonstrate that AllMatch achieves state-of-the-art performance and is capable of dealing with real-world challenges.

\bibliographystyle{named}
\bibliography{ijcai23}

\clearpage
\appendix

\section{More Quantitative Analysis}

\begin{figure}[t]
    \centering
    \subfigure[$\lambda_b$ on CIFAR-10-40.]{
        \includegraphics[width=0.22\textwidth]{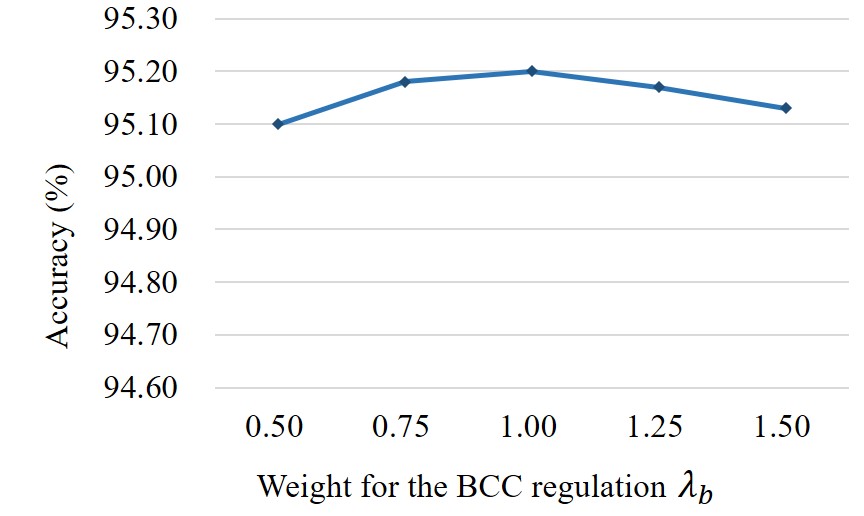}
        \label{weight_cifar10}
    }
    \subfigure[$\lambda_b$ on CIFAR-100-400.]{
        \includegraphics[width=0.22\textwidth]{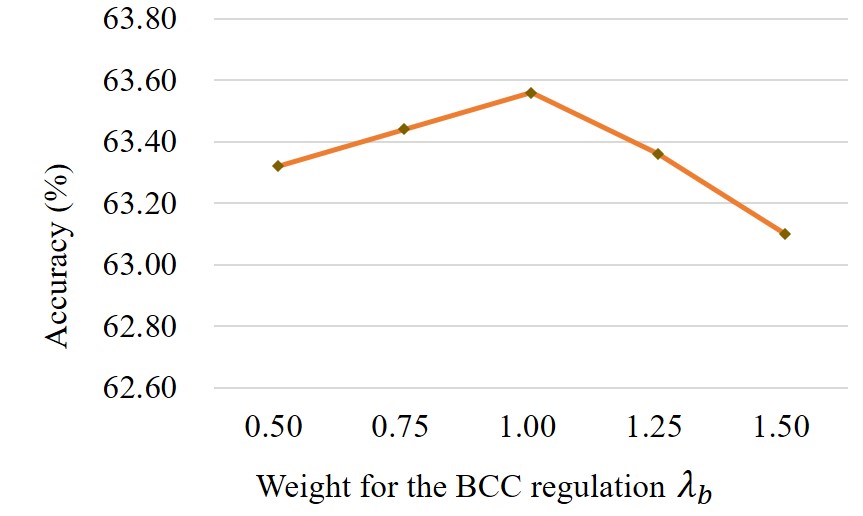}
        \label{weight_cifar100}
    }
    \caption{Grid search for the weight of BCC regulation $\lambda_b$.}
    \label{fig:weight}
\end{figure}

\subsection{Grid Search for the Weight of the BCC Regulation.}
The weight associated with the BCC regulation in the overall loss, denoted as $\lambda_b$, is systematically evaluated in Figure~\ref{fig:weight}. For both CIFAR-10 with 40 labeled samples and CIFAR-100 with 400 labeled samples, AllMatch achieves optimal performance when $\lambda_b$ is set to 1.0. Either larger or smaller values can result in slight performance degradation. Overall, AllMatch assigns equal importance to the supervision signals of pseudo-label and candidate-negative division, leveraging the latter to boost the potential within the top-k predictions of low-confidence pseudo-labels.

\begin{figure}[t]
    \centering
    \subfigure[$K$ on CIFAR-10-40.]{
        \includegraphics[width=0.22\textwidth]{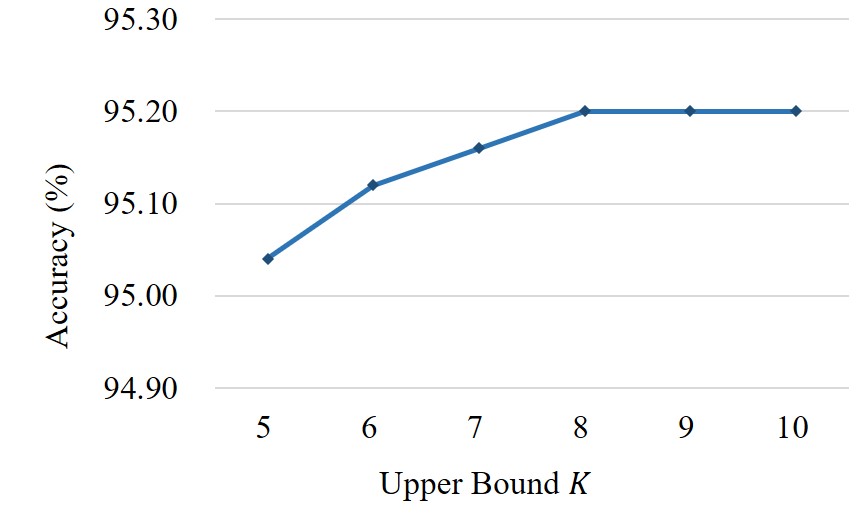}
        \label{k_cifar10}
    }
    \subfigure[$K$ on CIFAR-100-400.]{
        \includegraphics[width=0.22\textwidth]{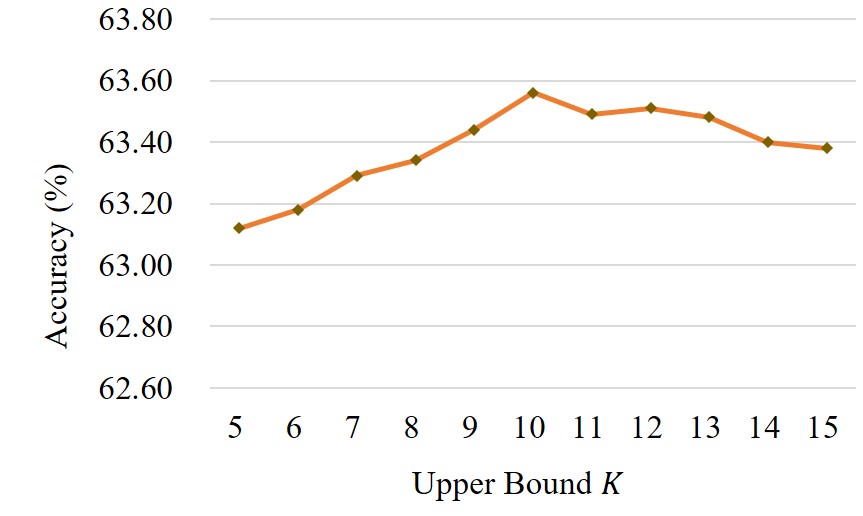}
        \label{k_cifar100}
    }
    \caption{Grid search for the upper bound of the number of candidate classes $K$.}
    \label{fig:k}
\end{figure}

\subsection{Grid Search for the Upper Bound of the Number of Candidate Classes.} As shown in Figure~\ref{fig:k}, we examine the upper bound of the number of candidate classes, denoted as $K$. In the case of CIFAR-10 with 40 labeled samples, the performance of AllMatch is largely unaffected by $K$ due to its effective distinction between candidate and negative classes through a comparison between local and global top-k confidence. Moreover, for CIFAR-100 with 400 labeled samples, the model achieves optimal performance when $K$ is set to 10. A smaller $K$ may exclude the ground truth from the candidate classes, leading to negative impacts. Additionally, a larger $K$ can result in a trivial division between candidate and negative classes, causing performance degradation. Considering the comparable learning difficulty, we set the upper bound $K$ to 10 for CIFAR-10/100, SVHN, and STL-10. However, ImageNet is known to be more challenging than the aforementioned datasets. Through grid search, we find that setting $K$ to 20 provides optimal results on ImageNet.

\begin{figure*}[t]
    \centering
    \subfigure[Class-average threshold.]{
        \includegraphics[width=0.22\textwidth]{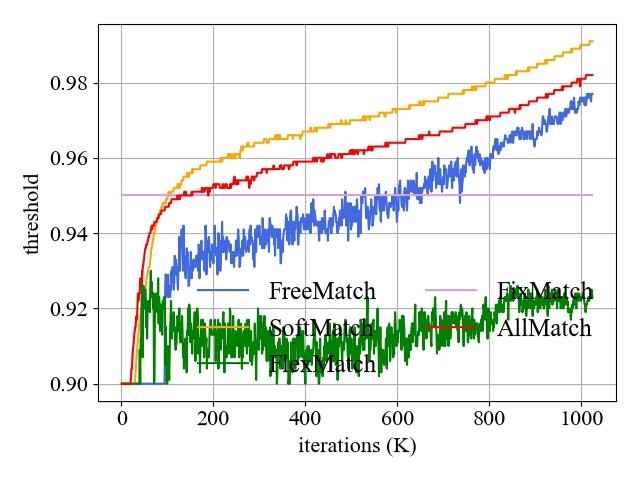}
        \label{threshold_cifar10_10}
    }
    \subfigure[Selected pseudo-label \textit{acc}.]{
        \includegraphics[width=0.22\textwidth]{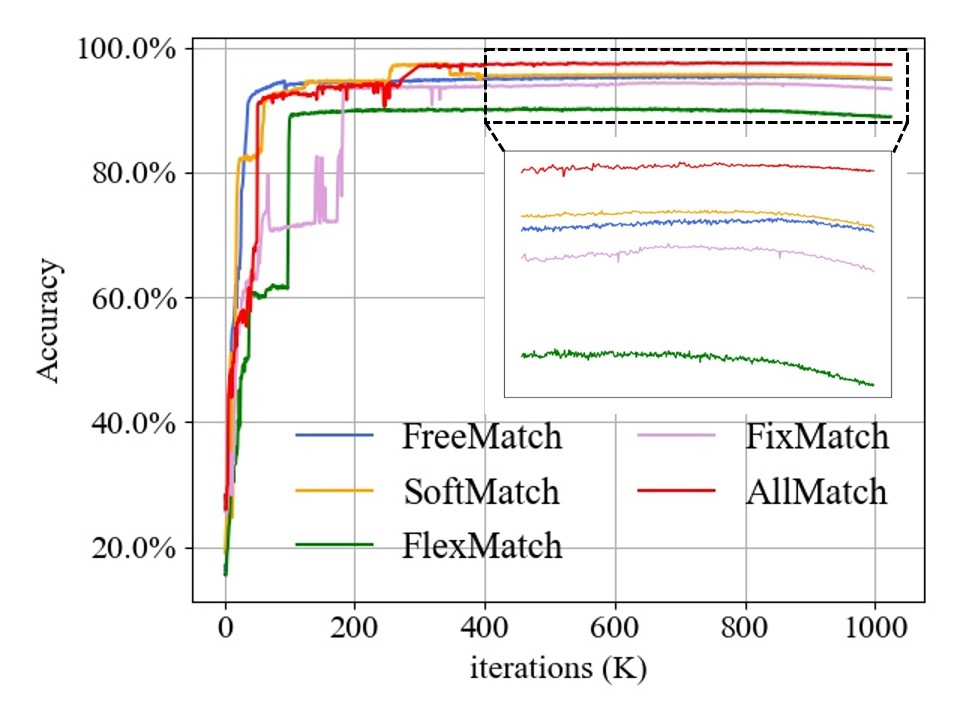}
        \label{select_pseudo_acc_cifar10_10}
    }
    \subfigure[Utilization ratio.]{
        \includegraphics[width=0.22\textwidth]{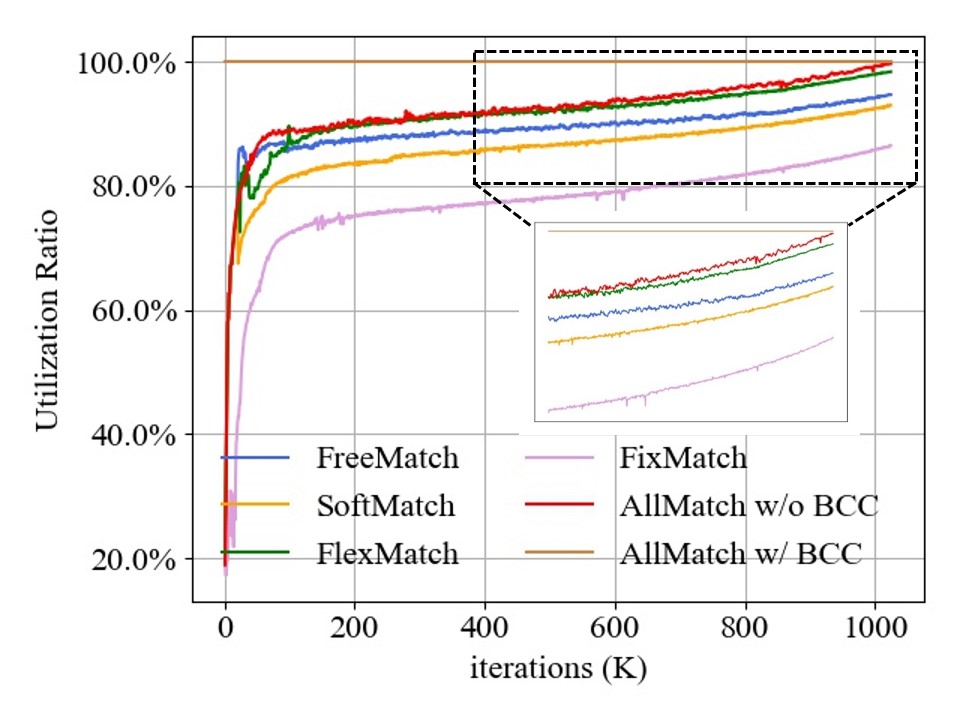}
        \label{utilization_ratio_cifar10_10}
    }
    \subfigure[Binary pseudo-label \textit{acc}.]{
        \includegraphics[width=0.22\textwidth]{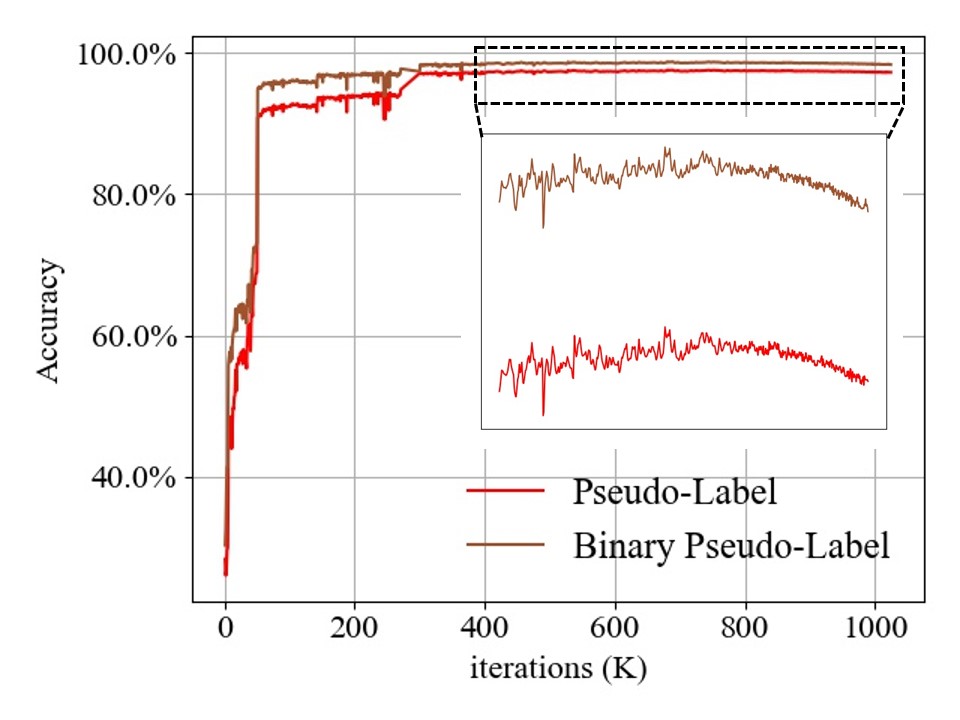}
        \label{binary_acc_cifar10_10}
    }
    \subfigure[Class-average threshold.]{
        \includegraphics[width=0.22\textwidth]{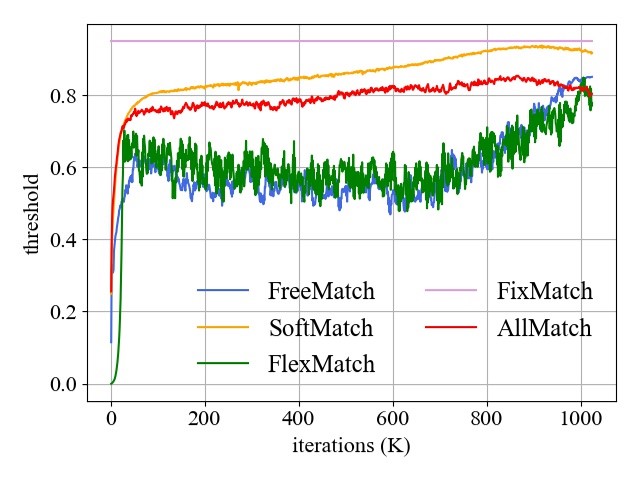}
        \label{threshold_cifar100}
    }
    \subfigure[Selected pseudo-label \textit{acc}.]{
        \includegraphics[width=0.22\textwidth]{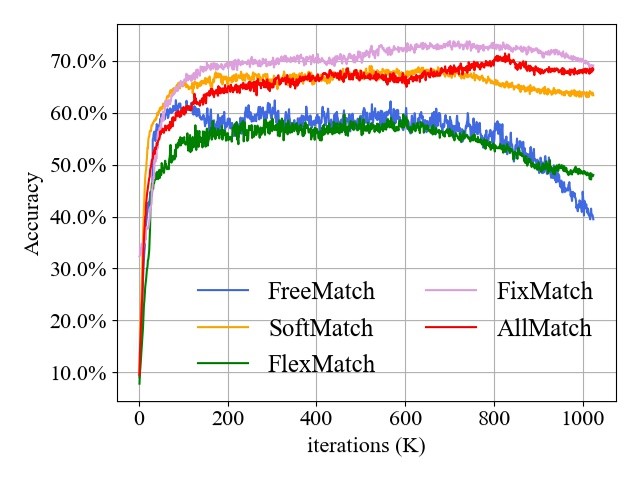}
        \label{select_pseudo_acc_cifar100}
    }
    \subfigure[Utilization ratio.]{
        \includegraphics[width=0.22\textwidth]{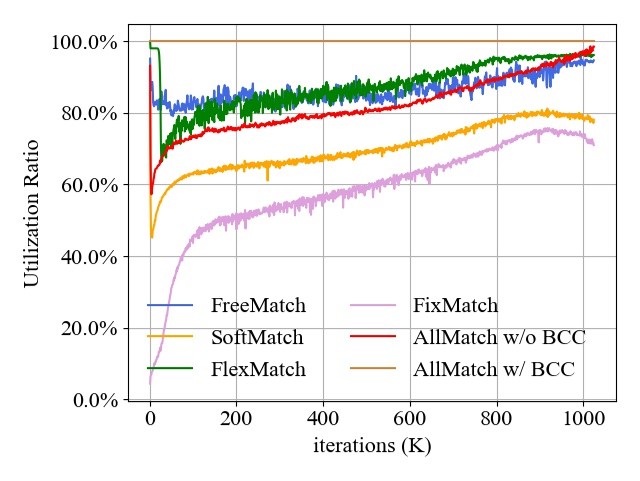}
        \label{utilization_ratio_cifar100}
    }
    \subfigure[Binary pseudo-label \textit{acc}.]{
        \includegraphics[width=0.22\textwidth]{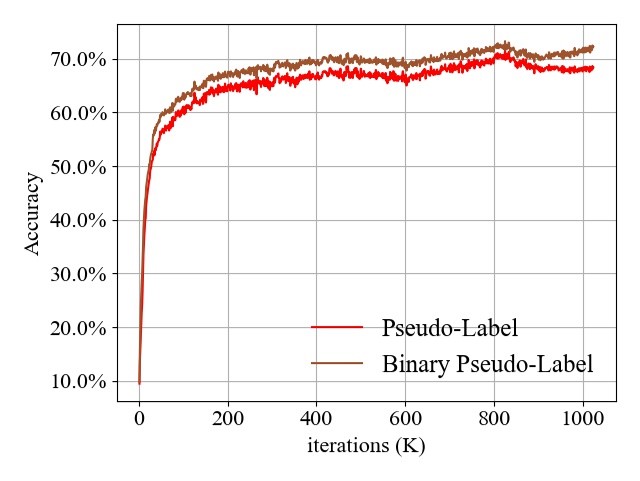}
        \label{binary_acc_cifar100}
    }
    \caption{How AllMatch performs on CIFAR-10 with 10 labels~(a-d) and CIFAR-100 with 400 labels~(e-h) compared to other methods. The threshold for CIFAR-10 with 10 labels is restricted within the range of [0.9, 1.0] to avoid overfitting noisy pseudo-labels in the early training stages. In the analysis of SoftMatch, we employ $\mu_t-\sigma_t$ as its threshold. Samples with a confidence lower than $\mu_t-\sigma_t$ are assigned negligible weights, essentially treated as if they were discarded. Consequently, we employ $\mu_t-\sigma_t$ as the class-average threshold of SoftMatch. Moreover, its selected pseudo-label \textit{acc} and utilization ratio for unlabeled data can be computed like other threshold-based algorithms. Here, $\mu_t$ and $\sigma_t$ denote the estimated \textit{mean} and \textit{standard deviation} of the overall confidence on unlabeled data.}
    \label{fig:indicators_cifar10_10_cifar100}
\end{figure*}

\begin{table*}[t]
    \centering
    \scriptsize
    \tabcolsep=0.08cm
    \begin{tabular}{c|ccc|ccc}
        \toprule
        Dataset & \multicolumn{3}{c}{CIFAR-10-LT} \vline & \multicolumn{3}{c}{CIFAR-100-LT} \\
        \midrule
        $\gamma$ & 50 & 100 & 150 & 20 & 50 & 100 \\
        \midrule
        FixMatch + ABC~(baseline) & 85.81{\tiny±0.29} & 81.12{\tiny±0.55} & 77.42{\tiny±0.91} & 53.10{\tiny±0.88} & 45.06{\tiny±0.68} & 41.33{\tiny±1.04}\\
        FlexMatch + ABC & 85.99{\tiny±0.22} & 81.05{\tiny±0.47} & 77.04{\tiny±1.05} & 53.33{\tiny±0.53} & 45.17{\tiny±0.49} & 41.08{\tiny±0.77}\\
        SoftMatch + ABC & \underline{86.19{\tiny±0.17}} & \underline{81.49{\tiny±0.29}}& \underline{77.62{\tiny±0.77}} & \underline{53.89{\tiny±0.49}} & \underline{45.82{\tiny±0.38}} & \underline{41.79{\tiny±0.73}}\\
        FreeMatch + ABC & 85.49{\tiny±0.17} & 81.01{\tiny±0.39} & 77.33{\tiny±0.82} & 53.62{\tiny±0.39} & 45.44{\tiny±0.61} & 41.48{\tiny±0.53}\\
        \midrule
        AllMatch + ABC & \textbf{86.52{\tiny±0.14}} & \textbf{81.83{\tiny±0.33}} & \textbf{78.02{\tiny±0.52}} & \textbf{54.19{\tiny±0.46}} & \textbf{45.91{\tiny±0.46}} & \textbf{41.96{\tiny±0.67}}\\
        \bottomrule
        \end{tabular}
    \caption{Performance of the combination of ABC with existing balanced SSL algorithms on CIFAR-10-LT and CIFAR-100-LT. ABC trains an auxiliary balanced classifier for FixMatch, and thus \textit{FixMatch + ABC} serves as the baseline model in the experiment.}
    \label{tab:abc_imbalance_ssl}
\end{table*}

\subsection{Learning Process Visualization} 

To gain further insights into AllMatch, we compare its learning process with previous algorithms on four benchmarks: CIFAR-10 with 40 labels, STL-10 with 40 labels, CIFAR-10 with 10 labels, and CIFAR-100 with 400 labels. The former two are presented in Figure 4 of the main paper and the latter two are provided in Figure~\ref{fig:indicators_cifar10_10_cifar100} of the Appendix. Among the involved SSL algorithms, SoftMatch serves as a weight-based method, varying from other threshold-based algorithms. In the following parts, we will first describe the analysis regarding SoftMatch in detail and subsequently provide the revealed findings about the learning process of AllMatch.

SoftMatch models sample weights by a dynamic Gaussian function, where the \textit{mean} $\mu_t$ and \textit{standard deviation} $\sigma_t$ are evaluated by predictions on unlabeled data. The weight for unlabeled data $u$, denoted as $\lambda(u)$, is defined as follows.
\begin{align}
   \lambda(u) = e^{-\frac{\text{min}(c - \mu_t, 0)^2}{2 \times (\sigma_t / n)^2}}
   \label{equation: softmatch}
\end{align}
Here, $c$ denotes the prediction confidence of sample $u$ and $n$ is set to $2$ in SoftMatch to adjust the \textit{standard deviation}. While each unlabeled sample receives a positive weight, samples with confidence significantly lower than $\mu_t$ are assigned close-to-zero weights, thus having minimal impact on the overall loss. Consequently, in this paper, we distinguish between the implementation and analysis of SoftMatch. Specifically, the implementation follows the sample weight scheme mentioned in Equation~(\ref{equation: softmatch}). Besides, in the analysis part, \textbf{we consider unlabeled samples with confidence lower than $\mu_t-\sigma_t$~($\lambda(u) < e^{-2}$) as dropped samples}, thereby approximating SoftMatch as a threshold-based method. Based on the above analysis, the class-average threshold for SoftMatch is defined as $\mu_t - \sigma_t$, and its selected pseudo-label \textit{acc} and utilization ratio for unlabeled data can be computed like other threshold-based algorithms.

Figure~\ref{fig:indicators_cifar10_10_cifar100} presents the evolving process of numerous training indicators on CIFAR-10 with 10 labels and CIFAR-100 with 400 labels. AllMatch demonstrates several commonalities between the two datasets. Firstly, the threshold exhibits the expected behavior, beginning with a small value and steadily increasing thereafter. Secondly, AllMatch shows a smooth threshold evolution on both benchmarks, thus providing a better estimation of the learning status when compared to previous models. Thirdly, while prior methods suffer from the degradation of pseudo-label accuracy in the later training stages, AllMatch effectively alleviates this issue. Lastly, the binary pseudo-label accuracy consistently outperforms the pseudo-label accuracy, indicating that the BCC regulation effectively identifies the candidate class for all unlabeled data.

In addition to the aforementioned commonalities, there are also some differences in the performance of AllMatch on different benchmarks. For CIFAR-10 with 10 labels, AllMatch not only achieves improved pseudo-label accuracy but also utilizes the unlabeled data more effectively, highlighting its significant advantages over existing algorithms, especially when labeled samples are extremely limited. On the other hand, for CIFAR-100 with 400 labels, FixMatch achieves the optimal pseudo-label accuracy by leveraging a high constant threshold throughout training. In comparison, AllMatch achieves comparable pseudo-label accuracy while substantially improving the utilization ratio of the unlabeled set, indicating a better trade-off between the utilization of unlabeled data and pseudo-label accuracy.

\begin{figure*}[t]
    \centering
    \subfigure[AllMatch (unlabeled set / test set).]{
        \includegraphics[width=0.46\textwidth]{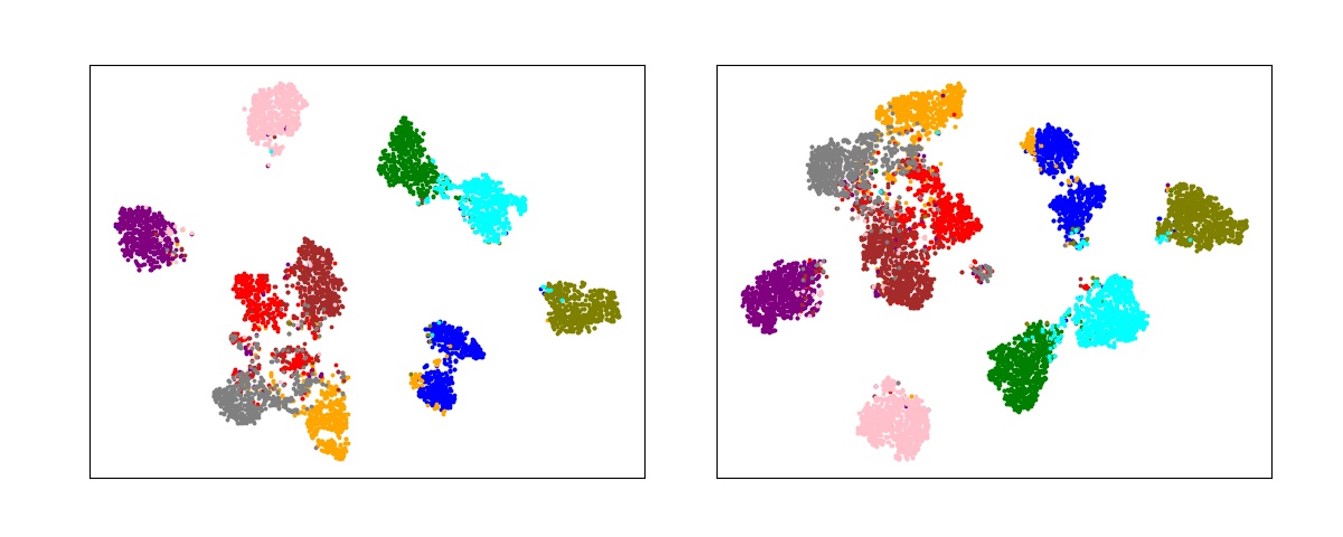}
        \label{AllMatch_tsne}
    }
    \subfigure[SoftMatch (unlabeled set / test set).]{
        \includegraphics[width=0.46\textwidth]{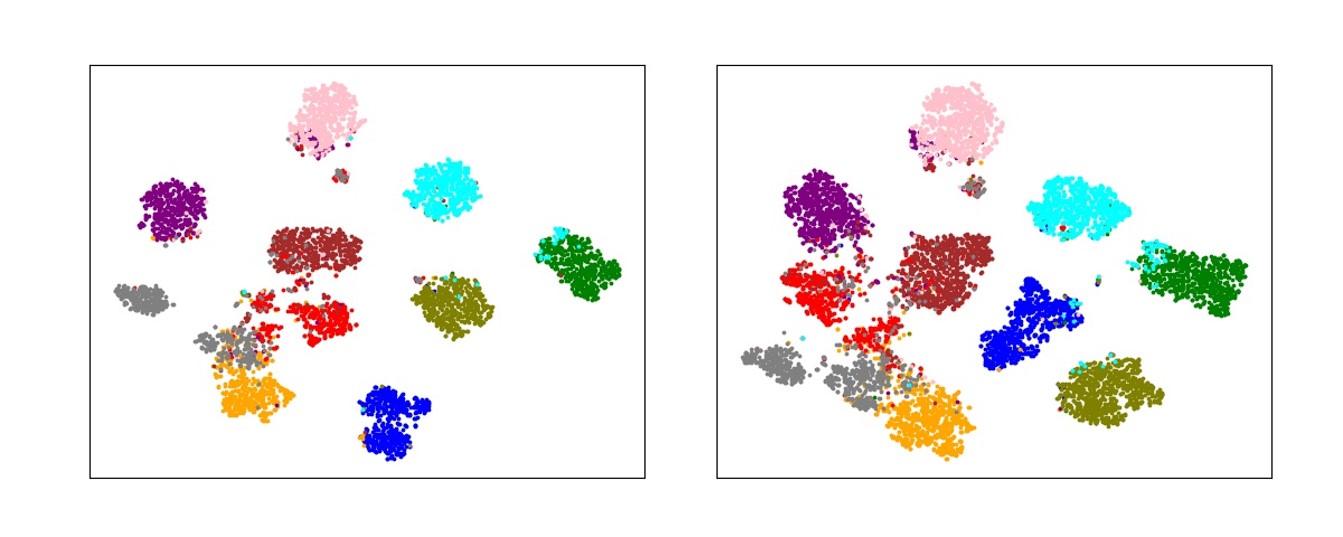}
        \label{SoftMatch_tsne}
    }
    \subfigure[FreeMatch (unlabeled set / test set).]{
        \includegraphics[width=0.46\textwidth]{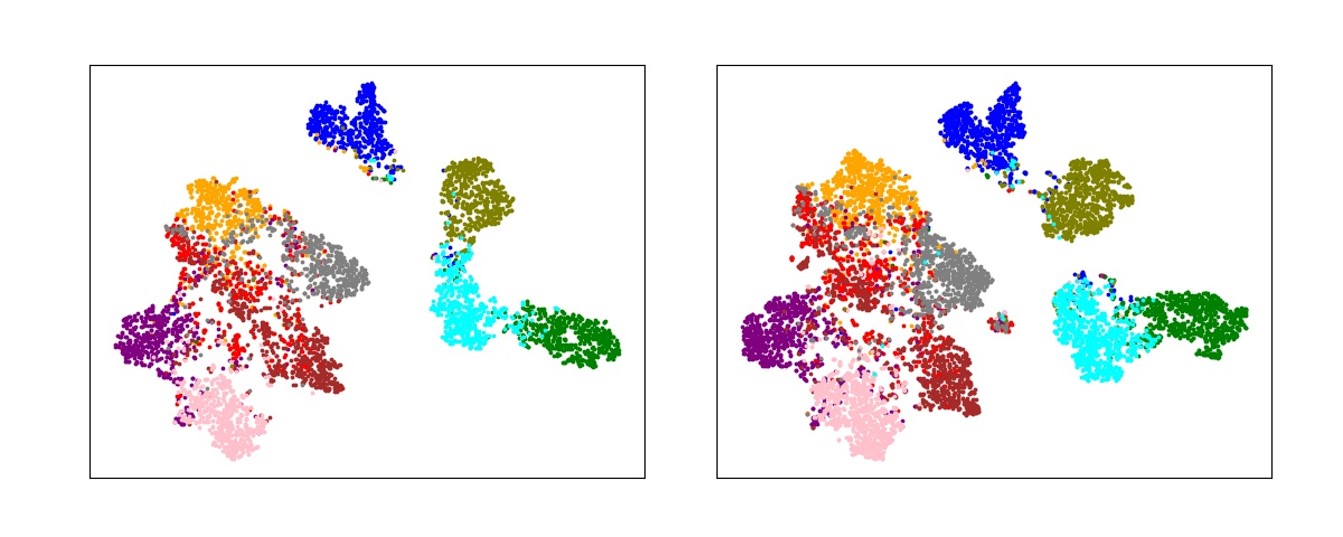}
        \label{FreeMatch_tsne}
    }
    \subfigure[FlexMatch (unlabeled set / test set).]{
        \includegraphics[width=0.46\textwidth]{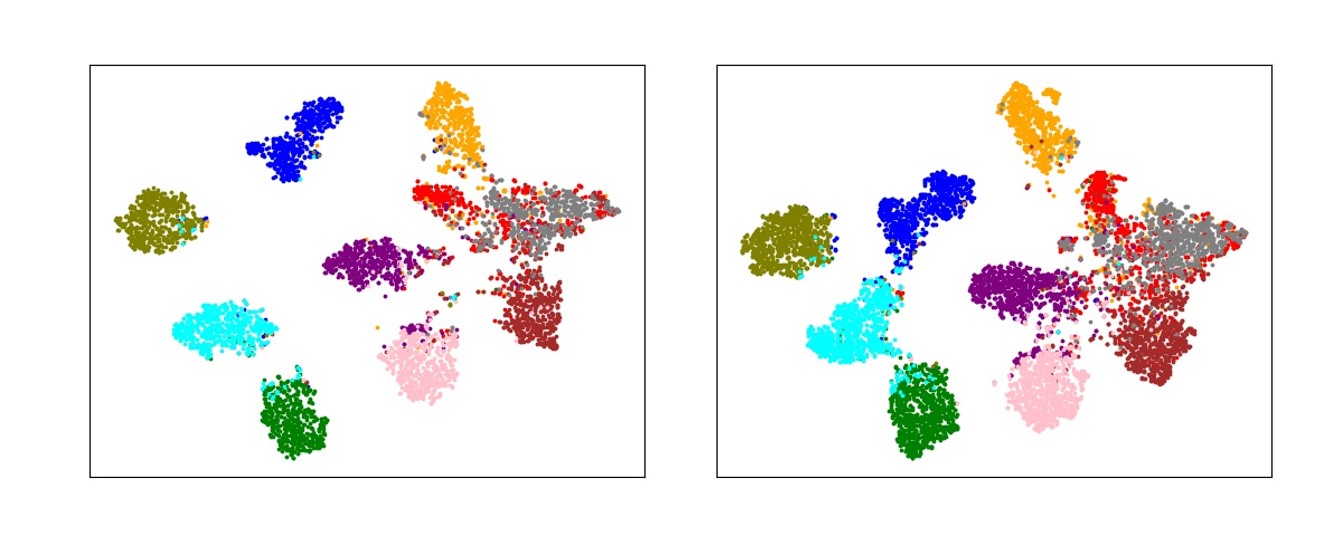}
        \label{FlexMatch_tsne}
    }
    \caption{Feature visualization for different SSL algorithms on STL-10 with 40 labeled samples.}
    \label{fig:tsne}
\end{figure*}

\begin{table}[t]
    \scriptsize
    \centering
    \tabcolsep=0.08cm
    \begin{tabular}{c|c c c c c}
    \toprule
    Datasets & CIFAR-10 & CIFAR-100 & SVHN & STL-10 & ImageNet \\
    \midrule
    Backbone & WRN-28-2 & WRN-28-8 & WRN-28-2 & WRN-37-2 & ResNet-50 \\
    \midrule
    Weight Decay & 5e-4 & 1e-3 & 5e-4 & 5e-4 & 3e-4 \\
    \midrule
    $B_L$ / $B_U$ & \multicolumn{4}{c}{64 / 448} & 128 / 128 \\
    \midrule
    LR / Scheduler & \multicolumn{5}{c}{0.03 /  $\eta=\eta_0 cos(\frac{7 \pi k}{16K})$ } \\
    \midrule
    SGD Momentum & \multicolumn{5}{c}{0.9} \\
    \midrule
    m / Model EMA & \multicolumn{5}{c}{0.999 / 0.999} \\ 
    \midrule
    K & \multicolumn{4}{c}{10} & 20 \\
    \midrule
    $\lambda_u$ / $\lambda_b$ & \multicolumn{5}{c}{1.0 / 1.0} \\
    \midrule
    Weak/Strong Aug & \multicolumn{5}{c}{Random Crop, Random Horizontal Flip / RandAugment}\\ 
    \bottomrule
    \end{tabular}
    \caption{Detailed training settings for balanced SSL.}
    \label{tab:param_balanced_ssl}
\end{table}

\begin{table}[t]
    \scriptsize
    \centering
    \begin{tabular}{c|c c}
    \toprule
    Datasets & CIFAR-10-LT & CIFAR-100-LT\\
    \midrule
    Backbone & \multicolumn{2}{c}{WRN-28-2} \\
    \midrule
    Weight Decay & \multicolumn{2}{c}{4e-5} \\
    \midrule
    $B_L$ / $B_U$ & \multicolumn{2}{c}{64 / 128}\\
    \midrule
    LR / Scheduler & \multicolumn{2}{c}{2e-3 /  $\eta=\eta_0 cos(\frac{7 \pi k}{16K})$ } \\
    \midrule
    Optimizer & \multicolumn{2}{c}{Adam} \\
    \midrule
    m / Model EMA & \multicolumn{2}{c}{0.999 / 0.999} \\ 
    \midrule
    K & \multicolumn{2}{c}{10}\\
    \midrule
    $\lambda_u$ / $\lambda_b$ & \multicolumn{2}{c}{1.0 / 1.0} \\
    \midrule
    Weak/Strong Aug & \multicolumn{2}{c}{Random Crop, Random Horizontal Flip / RandAugment}\\ 
    \bottomrule
    \end{tabular}
    \caption{Detailed training settings for imbalanced SSL.}
    \label{tab:param_imbalanced_ssl}
\end{table}

\subsection{T-SNE Visualization} 
To gain an intuitive understanding of AllMatch, we employ T-SNE to plot the high-dimensional features of various SSL algorithms, including SoftMatch, FreeMatch, FlexMatch, and AllMatch, on STL-10 with 40 labeled samples, as depicted in Figure~\ref{fig:tsne}. Compared with FreeMatch and FlexMatch, AllMatch achieves more separable and tightly clustered features, indicating that the introduced CAT serves as a better indicator for the evolving learning status of the model. Furthermore, while SoftMatch maintains appropriate inter-class and intra-class distances, it overfits many erroneous pseudo-labels, exemplified by the excessive light blue points within the green cluster. In contrast, AllMatch effectively pushes ambiguous pseudo-labels towards the decision boundary, thus minimizing their impact on the classifier. Consequently, AllMatch extracts easy-to-distinguish features for the unlabeled samples, establishing a solid foundation for a robust classifier. 

\begin{figure*}[t]
    \centering
    \subfigure[AllMatch (unlabeled set / test set).]{
        \includegraphics[width=0.46\textwidth]{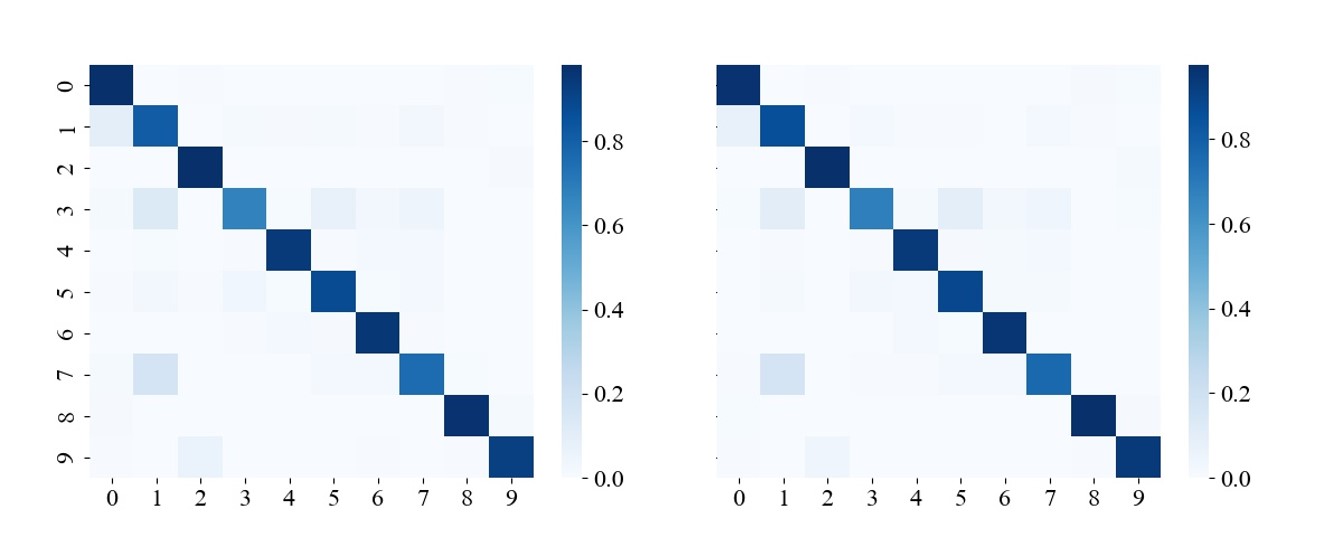}
        \label{AllMatch_confusion}
    }
    \subfigure[SoftMatch (unlabeled set / test set).]{
        \includegraphics[width=0.46\textwidth]{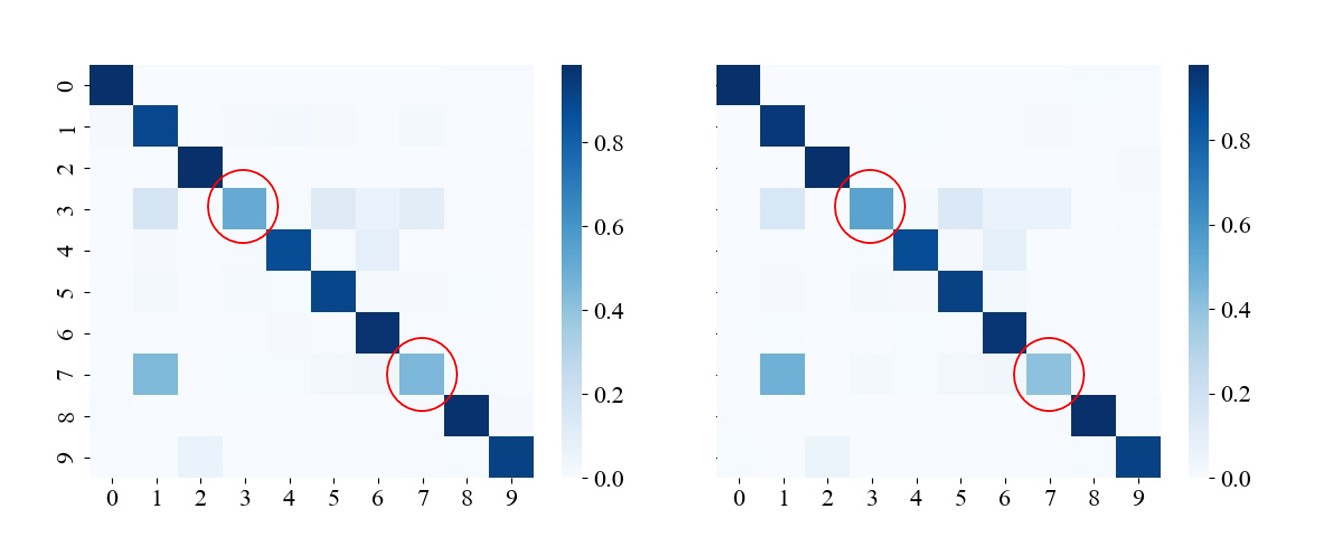}
        \label{SoftMatch_confusion}
    }
    \subfigure[FreeMatch (unlabeled set / test set).]{
        \includegraphics[width=0.46\textwidth]{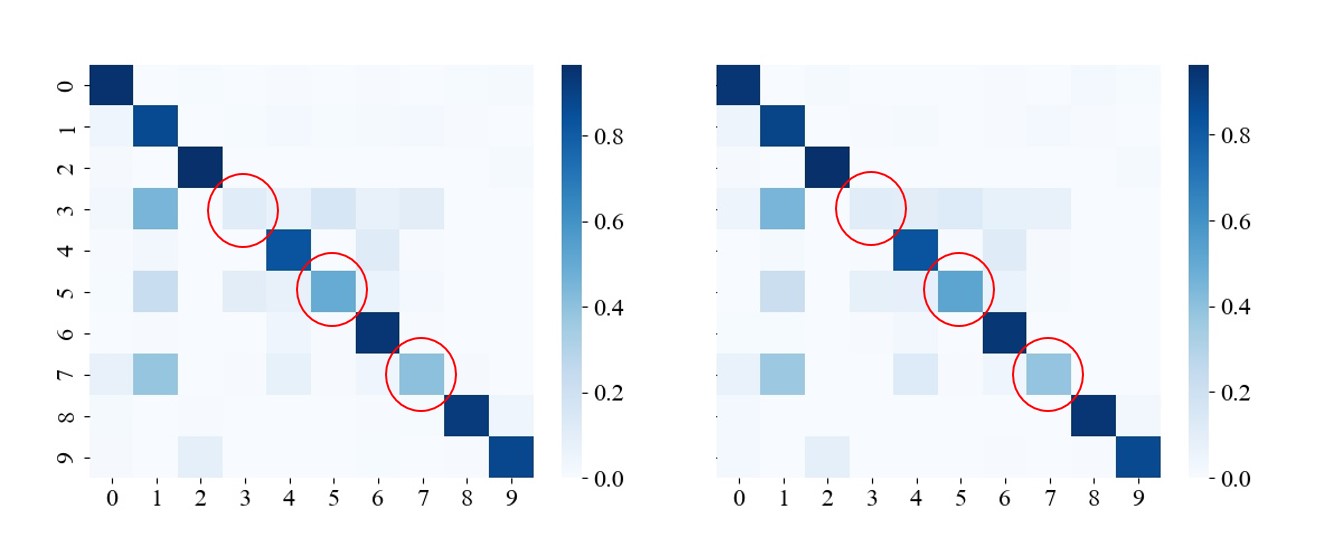}
        \label{FreeMatch_confusion}
    }
    \subfigure[FlexMatch (unlabeled set / test set).]{
        \includegraphics[width=0.46\textwidth]{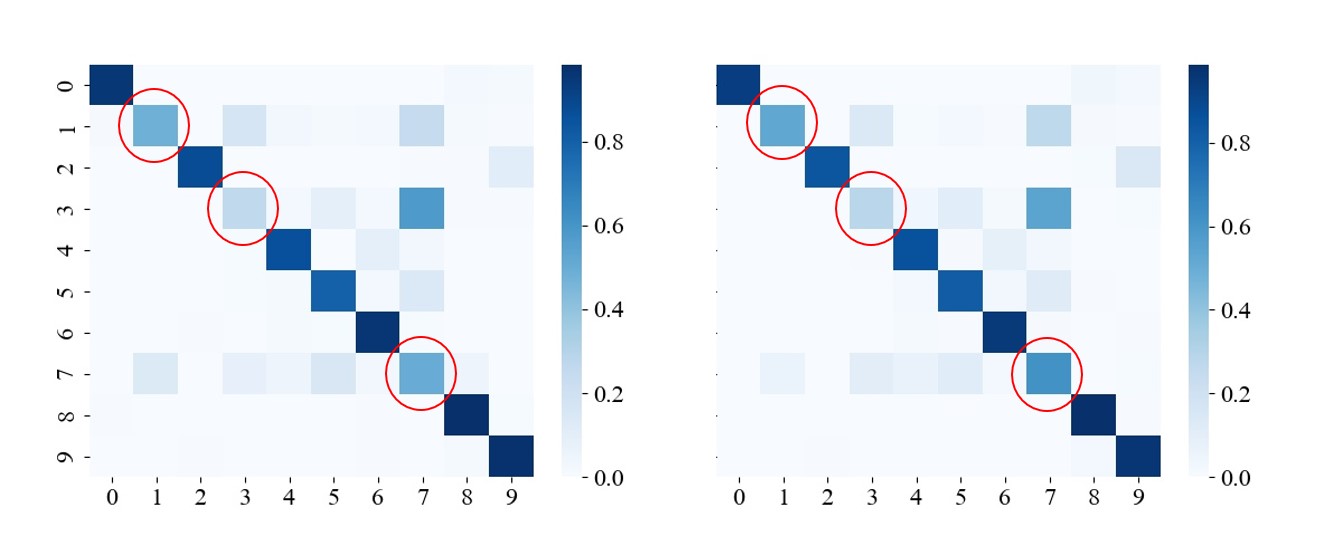}
        \label{FlexMatch_confusion}
    }
    \caption{Confusion matrices on STL-10 with 40 labeled samples. Classes with an accuracy below 0.7 are highlighted by red circles.}
    \label{fig:confusion}
\end{figure*}

\subsection{Confusion Matrix} In Figure~\ref{fig:confusion}, we provide the confusion matrices of several SSL algorithms when applied to STL-10 with 40 labeled samples: SoftMatch, FlexMatch, FreeMatch, and AllMatch. To emphasize the classes with an accuracy below 0.7, we denote them with red circles. The results presented in Figure~\ref{fig:confusion} suggest that previous models typically encounter challenges when recognizing samples in class 3, class 5, and class 7. Fortunately, AllMatch successfully mitigates this issue and promotes the accuracy of these three categories, which primarily stems from the accurate learning status estimation provided by CAT and the maximal utilization of the unlabeled set supported by BCC. Overall, AllMatch achieves enhanced class-wise accuracy compared to previous models, indicating its effectiveness. 

\section{Combination with Imbalanced Algorithms}
To explore the compatibility of the proposed method with existing imbalanced SSL algorithms, we examine the combination of ABC, which is the current state-of-the-art imbalanced SSL algorithm, with several balanced SSL algorithms on CIFAR-10-LT and CIFAR-100-LT. ABC uses Bernoulli masks to approximate class-balanced sampling, thus learning a balanced auxiliary classifier. As illustrated in Table~\ref{tab:abc_imbalance_ssl}, the combination of ABC and AllMatch consistently outperforms the baseline and other methods. Therefore, AllMatch can be jointly deployed with existing imbalanced SSL algorithms when encountering severe class imbalance.

\section{Detailed Training Settings}
For better reproduction, we present the detailed training settings for balanced and imbalanced SSL in Table~\ref{tab:param_balanced_ssl} and Table~\ref{tab:param_imbalanced_ssl}, respectively.

\end{document}